\pdfoutput=1
\documentclass[11pt]{article}

\usepackage[table, dvipsnames]{xcolor}
\usepackage{cite}
\usepackage{notoccite}
\usepackage{xspace}

\usepackage{naacl2021}

\usepackage{times}
\usepackage{latexsym}

\usepackage[T1]{fontenc}
\usepackage[utf8]{inputenc}

\usepackage{microtype}

\usepackage{cite}
\usepackage{notoccite}
\usepackage{amsmath,amssymb,amsfonts}
\usepackage{algorithmic}
\usepackage{graphicx}
\usepackage{changepage}
\usepackage{textcomp}
\usepackage{booktabs}
\usepackage{tabularx}
\usepackage{multirow}
\usepackage{url} %
\usepackage{lscape} 

\usepackage[inline]{enumitem}

\usepackage{xifthen}
\newcommand{\ifequals}[3]{\ifthenelse{\equal{#1}{#2}}{#3}{}}
\newcommand{\case}[2]{#1 #2} %
\newenvironment{switch}[1]{\renewcommand{\case}{\ifequals{#1}}}{}
\newcommand{\segstyle}[1]{\underline{\textbf{\emph{#1}}}}

\usepackage{ifthen}
\newcommand{\segment}[3]{
    \ifthenelse{1=0}{
        \begin{switch}{#1}
            \case{0}{\color{red}} %
            \case{1}{\color{orange}} %
            \case{2}{\color{blue}} %
            \case{3}{\color{green}} %
            
            \segstyle{* #2: #3}
            
        \end{switch}
        }{}
}

\newcommand{\cmnt}[1]{}
\newcommand{\todo}[1]{}

\newcommand{\codwoe}{\texttt{CODWOE}\xspace}
\newcommand{\defmod}{\texttt{DEFMOD}\xspace}
\newcommand{\revdict}{\texttt{REVDICT}\xspace}
\newcommand{\sgns}{\texttt{sgns}\xspace}
\newcommand{\electra}{\texttt{electra}\xspace}
\newcommand{\chr}{\texttt{char}\xspace}
\newcommand{\none}{} %
\newcommand{\allvec}{\texttt{allvec}\xspace}

\newcommand{\gru}{\texttt{gru}\xspace}
\newcommand{\lstm}{\texttt{lstm}\xspace}
\newcommand{\codwoerepo}{\url{https://github.com/TimotheeMickus/codwoe/}}
\newcommand{\irbnlprepo}{\url{https://github.com/dkorenci/codwoe-irb-nlp/}}

\newcommand{\mvsc}{MVR}
\newcommand{\sbleu}{BLEU}
\newcommand{\lbleu}{lBLEU}
\newcommand{\bst}[1]{\textbf{#1}} %

\newcommand{\besttrow}{\rowcolor[gray]{0.7}}
\newcommand{\submitrow}{\rowcolor[gray]{0.9}}

\title{IRB-NLP at SemEval-2022 Task 1: Exploring the Relationship Between Words
and Their Semantic Representations}

\author{Damir Korenčić\Thanks{\ Equal contribution.} \\
  Division of Electronics \\
  Ruđer Bošković Institute \\
  Zagreb, Croatia \\
  \texttt{damir.korencic@irb.hr} \\\And
  Ivan Grubišić\footnotemark[1] \\
  Division of Electronics \\
  Ruđer Bošković Institute \\
  Zagreb, Croatia \\
  \texttt{ivan.grubisic@irb.hr} \\}

\begin{document}
\maketitle
\begin{abstract}
What is the relation between a word and its description, or a word and its embedding? Both descriptions and embeddings are semantic representations of words. But, what information from the original word remains in these representations? Or more importantly, which information about a word do these two representations share? Definition Modeling and Reverse Dictionary are two opposite learning tasks that address these questions. The goal of the Definition Modeling task is to investigate the power of information laying inside a word embedding to express the meaning of the word in a humanly understandable way -- as a dictionary definition. Conversely, the Reverse Dictionary task explores the ability to predict word embeddings directly from its definition. In this paper, by tackling these two tasks, we are exploring the relationship between words and their semantic representations. We present our findings based on the descriptive, exploratory, and predictive data analysis conducted on the CODWOE dataset. We give a detailed overview of the systems that we designed for Definition Modeling and Reverse Dictionary tasks, and that achieved  top scores on SemEval-2022 CODWOE challenge in several subtasks.  We hope that our experimental results concerning the predictive models and the data analyses we provide will prove useful in future explorations of word representations and their relationships.

\end{abstract}

\section{Introduction}

The \textbf{COmparing Dictionaries and WOrd Embeddings} (\codwoe) task \cite{codwoe-paper} is aimed at explaining two different types of semantic descriptions of words: dictionary glosses and word embeddings. A \emph{dictionary gloss} is a brief textual explanation of a word and a \emph{word embedding} is a vector representation that captures the word's semantic and syntactic properties \cite{smith2020-wordrep}. 

In order to investigate the relationship between these two types of descriptions, two complementary subtracks were put together:
\begin{enumerate*}
    \item{\emph{Definition Modeling} (\defmod) track, where correct glosses need to be generated from word embedding vectors \cite{noraset-2017-dm-definition}; and}
    \item{\emph{Reverse Dictionary} (\revdict) track, where correct embedding vectors should be generated from dictionary glosses \cite{hill-2016-rd-learning-understand}.}
\end{enumerate*}
The datasets for both tracks cover five different languages: English (EN), Spanish (ES),  French (FR), Italian (IT), and Russian (RU).

The key challenge of the \codwoe task is that it needs to be performed without external data, which precludes the use of pretrained models and vectors. Additionally, the training dataset is relatively small in comparison to the datasets on which models are typically trained. 

Our strategy was to adapt an RNN-based decoder model \cite{noraset-2017-dm-definition} for the \defmod track, 
and to use a transformer-based encoder \cite{devlin_bert_2019} for 
the \revdict track. With the limited amount
of available data in mind, we hypothesized that models should not be large.
Therefore we aimed to limit the model complexity by reducing the number of parameters,
for example by using a subword tokenizer \cite{kudo_sentencepiece_2018},
which yields a smaller dictionary of optimized subword fragments. %
All of the models we used were built for a single language, 
and their structure and parameters were optimized either 
iteratively or by way of Bayesian hyperparameter optimization (BHO) \cite{snoek-2012-practical-bayesian-optimization-ml}.

We conducted data analyses of the \codwoe datasets 
and analyses of the developed machine learning models. We performed a statistical and visual analysis of the pretrained \codwoe embeddings, i.e., of their distributions and relationships.
\defmod analyses include an analysis of model performance factors
and a qualitative analysis of generated glosses.
In the \revdict predictive analysis, we investigate the impact of many different settings on models' performance defined in terms of distance and similarity scores between predicted and target vectors.

We show that our adaptation of the \defmod architecture \cite{noraset-2017-dm-definition}
can perform competitively and that the use of multiple word embeddings can 
clearly improve the generation of word glosses. 
For \revdict, we demonstrate that our approaches achieve top 
performance in terms of ranking,
which makes them suitable for information retrieval applications.
Our models perform competitively and our results on the \codwoe challenge
can be found in Table \ref{tab:results}. 
We make the code of our models and data analyses publicly available\footnote{\irbnlprepo}.

\begin{table}
\small
\caption{
Aggregated language-level ranks of our team for the \defmod (DM) and \revdict (RD) tracks (and the number of teams competing in a subtask).
}
\vspace{-2mm}
\centering
\begin{tabular}{llllll}
\hline
\textbf{TASK} & \textbf{EN} & \textbf{ES} & \textbf{FR} & \textbf{IT} & \textbf{RU} \\
\hline
DM-all      & 2 (9) & \bst{1} (7) & \bst{1} (6) & 5 (7) & 5 (6) \\
RD-sgns     & 3 (9) & \bst{1} (7) & \bst{1} (6) & \bst{1} (7) & 2 (6) \\
RD-char     & 4 (7) & 3 (5) & 4 (5) & 2 (6) & 2 (5) \\
RD-electra  & 5 (6) &   & 3 (4) &   & 3 (4) \\
\hline
\end{tabular}
\vspace{-3mm}
\label{tab:results}
\end{table}

\section{Background}

\subsection{Related Work}

\paragraph{Definition Modeling}

The Definition Modeling (\defmod) task, first introduced in \citealp{noraset-2017-dm-definition}, 
is focused on the prediction of dictionary word glosses from word embeddings.
\citet{noraset-2017-dm-definition} experimented on two English dictionaries
and proposed a successful architecture based on RNN.

Subsequent work on Definition Modeling focused on variations of the 
problem of prediction of a word gloss from the word sense.
These approaches consider gloss prediction based on sense-specific 
word embeddings \cite{gadetsky-2018-dm-conditional, kabiri-2020-dm-evaluating, zhu_multi-sense_2019}, 
and on a word-based context indicating the word sense \cite{gadetsky-2018-dm-conditional, mickus_mark_2019, bevilacqua-2020-dm-generationary, zhang-2020-dm-interpretability, yang-2020-dm-sememes}.
The proposed approaches are based either on 
RNNs \cite{gadetsky-2018-dm-conditional, kabiri-2020-dm-evaluating, zhu_multi-sense_2019, zhang-2020-dm-interpretability} 
or Transformers \cite{mickus_mark_2019, bevilacqua-2020-dm-generationary}.
All of the previous approaches rely on word embeddings
pre-trained on large corpora, most commonly word2vec \cite{mikolov-2013-embed-word2vec}.
Sense-aware approaches that take embeddings as input make use 
of either sense-aware word embeddings \cite{gadetsky-2018-dm-conditional, kabiri-2020-dm-evaluating}
or of decomposition of word embeddings into sense-specific vectors \cite{zhu_multi-sense_2019}.

The initially proposed architecture of \citet{noraset-2017-dm-definition} is often used as a baseline solution. 
The most commonly used measure of model performance is the BLEU \cite{papineni-2002-bleu} metric.
Although there is some overlap in used datasets, most experiments rely on a specific dataset. 
The reported model performances vary greatly. 
\citet{noraset-2017-dm-definition} report BLEU of $31$ and $23$, depending on the dictionary.
Subsequent experiments report, for the same approach, BLEU scores that range 
from as little as $11$ \cite{gadetsky-2018-dm-conditional} to as much as $60$ \cite{kabiri-2020-dm-evaluating}.
The variation can be great even for the same language and experimental setup \cite{kabiri-2020-dm-evaluating}.
The original approach of \citet{noraset-2017-dm-definition} remains competitive in the sense-aware setting, 
with the sense-aware approaches achieving BLEU increases that range between $1-2$ 
\cite{gadetsky-2018-dm-conditional, kabiri-2020-dm-evaluating, zhang-2020-dm-interpretability}
and $5-6$ \cite{yang-2020-dm-sememes, kabiri-2020-dm-evaluating, zhang-2020-dm-interpretability}, 
depending on the setting.

While we view the Definition Modeling primarily as a theoretically interesting
task, potential applications include explainability of word embeddings 
and automatic generation of dictionaries, which might be of interest
in low-resource settings.

\paragraph{Reverse Dictionary}

The Reverse Dictionary (\revdict) is a task of finding the right word when a word description is given \cite{dutoit-2002-rd-lexical, bilac-2004-rd-dictionary-search, zock-2004-rd-word}. It is the formulation of the tip-of-the-tongue problem (TOT) \cite{brown-1966-tip-of-the-tongue} that occurs during text synthesis. It is a condition in which a person knows a lot about the word, such as its meaning and origin, but is unable to recall it.
\revdict is a complex task. There are countless variations of input definitions that should lead to the same one-word concept. This complexity comes in part from the representation of the one-word concepts in the human mind. People tend to relate concepts on the conceptual and lexical level and form a highly connected network of abstractions \cite{zock-2004-rd-word}.

Therefore, a natural approach to solving \revdict is to form a semantic network with nodes (one-word concepts) and edges (associations) to search for the target word \cite{zock-2004-rd-word, thorat-2016-rd-implementing}.
\revdict can be realized directly by comparing the input definitions with all the definitions in the dictionary and returning the most similar ones, without taking into account any semantic or grammatical information \cite{el-2004-rd-wordnet}.
However, \revdict systems that include semantics give better results, such as in \citealp{mendez-2013-rd-wordnet} and \citealp{calvo-2016-rd-concept-blending} where words are represented as vectors in a semantic space.

Recent \revdict approaches utilize deep learning (DL) to map arbitrary-length definition phrases to the vector representation of the target word \cite{hill-2016-rd-learning-understand, qi-2020-rd-wantwords,  yan-2020-rd-bert, malekzadeh-2021-rd-predict}. The success of DL approaches indicates that \revdict can be solved implicitly, i.e. by directly learning from given data, and doesn't require an explicit injection of domain knowledge. According to this observation, the DL approach is a good choice for solving the \revdict task. 

\subsection{Dataset}

The \codwoe datasets \cite{codwoe-paper} cover five languages (EN, ES, FR, IT, RU) and are derived from the Dbnary lexical data\footnote{\url{http://kaiko.getalp.org/about-dbnary/}}. Each data point corresponds to a single word and contains word embedding vectors and the word gloss. Three types of embedding are used, labeled as \sgns (pretrained word2vec), \electra (contextual pretrained embeddings) and \chr (character-based embeddings). Pretrained embeddings are based on large corpora containing approximately 1B tokens.

Each dataset is divided into three sections: training, validation (development), and test. Datasets for training and validation have 43.608 and 6.375 samples, respectively. Each track also has a separate set of test data. The \defmod test dataset has 6.221 samples while the \revdict has 6.208 samples.

More detailed statistics and analyses of the dataset can be found in the Appendices, including the gloss statistics (Table \ref{tab:orig-gloss-stats}) and embedding vector statistics (Table \ref{tab:train-data-stats}).
Descriptive analysis of the embedding vectors shows large variation in values
that depend on a language and an embedding type (Figures \ref{fig:dist-train-vec-full} and \ref{fig:dist-train-vec}). Additionally, an exploratory analysis showed that the embeddings for different languages are easily separable (Figures \ref{fig:dist-vec-2e-5l-2D} and \ref{fig:dist-vec-3e-3l-2D}).
Interestingly, patterns of vector-based word similarity seem to differ significantly across embedding types, and in this regard there are no visible relations between different embeddings (Figure \ref{fig:dist-vec-3e-3l-2D-electra-clusters}).

\section{System overview}
\label{sect:sysoverview}

Both the \defmod and the \revdict models rely on unigram subword 
tokenizers \cite{kudo-subword-2018} trained on glosses from the train datasets.

\subsection{Definition Modeling}
\label{sect:defmodoverview}

Our approach to the challenging task of Definition Modeling on a limited dataset 
consists of preprocessing the input data, extracting the semantic information from the dataset, and controlling the model size and complexity.

The inspection of the learning data revealed that the gloss texts are often long since they
consist of several alternative definitions. We opted to include only one definition per learning example. Our intuition is that this approach, also taken in \cite{noraset-2017-dm-definition}, alleviates the learning problem by inducing the model to learn shorter and atomic definitions. 
The approach should also reduce noise (since the number of alternative definitions in a gloss is arbitrary).

The inspection of glosses also revealed the presence of lexicographic labels that precede the gloss definitions. These labels, present for all languages except English, 
convey data about, for example, word semantics (ex. geography, history) 
or temporal category (ex. archaic).
We chose to remove these labels since they introduce noise (the presence and the amount of labels
appears arbitrary), increase the dictionary size, and thus make the learning problem harder.

To construct the dictionary we use the unigram subword tokenizer \cite{kudo-subword-2018} 
implemented as part of the SentencePiece tool \cite{kudo_sentencepiece_2018}.
The reasons for using the subword tokenization were the expected improvement in performance
for low-resource tasks \cite{kudo-subword-2018} and the reduction in
the number of model parameters corresponding to token embeddings.

Since we opted for a deep learning model depending on token embeddings, 
we initialized the token embeddings with GloVe vectors \cite{pennington_glove_2014}
trained on the dataset of normalized and cleaned atomic glosses.
To demonstrate that the GloVe vectors capture a degree of word semantics, 
we aggregated the vectors on a gloss level using tf-idf weighting. 
Then we inspected, for each ``target'' gloss from a sample of English glosses, 
other dataset glosses ordered by cosine similarity to the target.
This revealed that GloVe similarity corresponds to the similarity in gloss meaning. 
Additionally, we found that the models
initialized with GloVe vectors achieve a lower final loss.

\paragraph{Machine learning model}
We decided to use an adaptation of the RNN-based model of \cite{noraset-2017-dm-definition}, 
that proved competitive in a number of experimental settings.
In the context of the \defmod task, the model takes as input one or more 
word embeddings (\sgns, \electra or \chr) and produces a gloss (a sequence of tokens)
that should correspond to the word's correct gloss.

From the input embeddings, we form two vectors, the \emph{seed} vector $s$ that 
is used to initialize the RNN, and the \emph{context} vector $c$. 
For both the seed and the context vectors we consider using 
a single embedding, concatenation of embeddings, and a 
nonlinear transformation of the concatenation.
At each position in the sequence the context vector is passed as input, 
together with the RNN's output, to the special GRU-like gated cell \cite{noraset-2017-dm-definition}.
The output of the gated cell is then transformed (via linear transformation and softmax activation) 
to produce token-level probabilities.
The gated cell can learn to effectively combine the semantic context with the RNN-level features
in guiding the generation process \cite{noraset-2017-dm-definition}.
The network architecture we use is labeled as S+G in \citet{noraset-2017-dm-definition}. 
The described model performs conditional generation of 
tokens in a sequence, which is a standard approach in RNN-based language modeling.
The probability of a gloss $g$ is factorized under the assumption that 
each token $g_i$ depends on the previous tokens, the seed embbedding $s$, and the context $c$ :
$$
p(g|s,c) = \prod\limits_{i=1}^{|g|} p(g_i|g_{0:i-1},s,c)
$$

In \cite{noraset-2017-dm-definition}, the context is equal to the seed, 
i.e., the input word embedding. 
In our case, both the seed and the context can either be a single embedding or 
a function of multiple embeddings. 
This approach enables us to leverage the information from several word embeddings in a flexible way.
For example, \sgns embeddings can be used as a seed 
while the context can be formed by passing all the embeddings through a multilayer perceptron.
Another important difference is that we use the unigram subword tokenization \cite{kudo-subword-2018}. 
Finally, we experiment with using both LSTM and GRU as the network's RNN components.

\subsection{Reverse Dictionary}
\label{sect:system-overview-rd}

We approach \revdict as a supervised vector regression task and employ an end-to-end deep learning solution. Our model is based on a transformer architecture \cite{vaswani-2017-attention-transformer} used as a definition sentence encoder, and a fully connected feed-forward network used as an output regression module.

The transformer is used to produce useful representations from given inputs, where the inputs are tokenized definition sentences.
For each subword token in the input sequence, the transformer gives a representation in the form of a vector. 
Our \revdict systems implement three different approaches for aggregating the output vectors produced by the transformer:
\begin{enumerate*}
    \item{\emph{sum}, where we sum the representations given for each token in the input sequence;}
    \item{\emph{average}, where we average the representations given for each token in the input sequence; and}
    \item{\emph{eos}, where we use only the representation of the last token in the input sequence, i.e. end-of-sequence (eos) token.}
\end{enumerate*}
The output module further transforms these representations into word embedding vectors.

Additionally, we utilize a multi-task learning \cite{caruana-1997-multitask, ruder-2017-multitask-overview} approach. To support multi-task learning, we implemented multiple output regression modules that simultaneously predict different types of embedding vectors from the same representations produced by a single encoder. Multi-task learning is used during the model training phase and only output from one output module makes final predictions.
The motivation for using a multi-task learning approach is to benefit from inductive transfer between tasks that could improve the results of predicting a single task \cite{caruana-1997-multitask}.

\section{Model Selection and Experimental Setup}

In this section we describe the technical details of data preprocessing and model selection
that comprise our methods of constructing the \defmod and \revdict models.
The conceptual description of the methods is given in Section \ref{sect:sysoverview}.

\subsection{Definition Modeling}
\label{sect:defmod-experim}

Our choices regarding the technical details of data preprocessing and model construction were guided by what we will call \emph{development experiments}. These experiments consisted of training the model on the train 
set, and observing both the final development set loss and the quality of the produced glosses.

Output gloss quality was assessed using a separate ``trial'' dataset - a small dataset of $200$ items provided by the organizers, containing gloss information consisting of the embedding vector, the original word, and the gloss text.
The assessment was performed for English glosses only and aimed to assess 
the quality of the generated text, and the similarity of the output and the original glosses. A choice was deemed an improvement if it led to the improvement of development loss and either improved the generated glosses or caused no degradation in gloss quality. 
The development of the final algorithm was performed iteratively and heuristically. However, the overall improvement over the iterations is confirmed by the results of the test set evaluations.

\paragraph{Dataset transformation}
The transformation of the original dataset is performed by creating unambiguous training examples and removing the uninformative data that makes
the problem harder.

In the original dataset a gloss definition often consists of several 
equivalent but differently phrased definitions.
We divided the dictionary glosses into atomic definitions by splitting the 
text strings around the ``;'' character. This heuristic was motivated by gloss sample analysis 
and the inspection of a sample of atomic glosses revealed that it works in the majority of cases.
Each atomic gloss in the new dataset was paired with all the embedding vectors of the original gloss.

In order to remove lexicographical labels from the beginning of the glosses' text, simple language-specific regular expressions and removal rules were formed based on gloss sample analysis.
This approach proved to be effective for a large majority of glosses.

To perform further normalization we additionally lowercased 
all the glosses and removed the punctuation from the end of texts.
The code used to preprocess the original dataset, the new dataset, and the transformation log can be found in the code repository. 
We note that both the SentencePiece dictionary and the GloVe vectors used for \defmod are derived from the transformed dataset.
The statistics of the transformed glosses are presented in Table \ref{tab:modified-gloss-stats}

\paragraph{Dictionary}
We used the unigram subword tokenizer \cite{kudo-subword-2018} available 
as part of the SentencePiece tool \cite{kudo_sentencepiece_2018}.
and trained it using the default parameters. 
Experiments in \citet{gowda2020} suggest that a vocabulary of $8000$ subwords
is a good default choice for several languages in the case of machine translation.
Additionally, our \emph{development experiments} showed that English models 
using a vocabulary of $8000$ subwords are superior to $10000$ subword models.
Therefore we decided to set the number of unigram tokens to  $8000$ 
in case of English, and to $8500$ in case of other, highly inflected languages expected to have a higher number of distinct suffixes.

\paragraph{Pretrained token embeddings}
GloVe embeddings \cite{pennington_glove_2014} of the subword tokens, 
introduced to initialize the tokens with corpus-level semantic information,
were constructed as follows. The model was trained on the set of transformed glosses, and the embedding size was fixed to $256$ (the size of the gloss embeddings).
The number of training iterations was set to $50$, the ``cutoff'' parameter
$x_{max}$ was set to $10$, while all the other parameters retained their default values.
No frequency-based vocabulary pruning was performed.

\paragraph{Machine learning model}

We fixed the maximum sequence length of the RNN models to $64$ subword 
tokens. Our intuition is that this alleviates the learning problem
and could lead to models focused on generating shorter but more correct glosses.

The models were optimized using the AdamW algorithm \cite{loshchilov2017adamw}
and the standard categorical cross-entropy loss.
The training process was stopped after a fixed number of epochs, 
or if the best solution did not improve by more than $0.1\%$ over $10$ epochs.
During inference, the optimal solution was constructed using the 
beam search algorithm implementation provided by the competition organizers\footnote{\codwoerepo}.

We iteratively improved the models using the described \emph{development experiments}, 
i.e., relying on the development set loss and analysis of model glosses produced for the trial dataset.
We experimented with several architectural elements and hyperparameters:
the formulation of the seed (RNN init. value) and context (gate input) of the network, 
RNN cell type, dropout, learning rate (LR) and LR scheduler, and the number of training epochs.

The most successful variant is constructed by using the concatenation of all the gloss embeddings as the context and the \sgns embedding as the seed.
This variant uses input dropout of $0.1$ and network dropout of $0.3$.
The input dropout is applied to the seed and context vectors, as well as to the word embeddings.
The network dropout is applied to the output of the RNN (final layer) and to the output of the gate cell.
The chosen learning rate  is $0.001$, and the ``plateau'' LR scheduler is used --
LR is multiplied by $0.1$ if there is no improvement over $5$ epochs.

For the context vector, we tried single embeddings and the combined embeddings merged via a multilayer
perceptron. Both variants proved inferior to the concatenation of all vectors.
The merged seed vector proved no different from the single embedding seed, so we opted for the simpler solution.
Both the development experiments and the results showed no difference between the LSTM and the GRU cell.

Analysis of errors revealed that models sometimes produce a deformed
output (very short or non-alphabetic string), and that this
almost never occurs simultaneously for two distinct models.
Therefore a way of heuristic model improvement is to combine it with another \emph{fallback} model to be used in case of deformed outputs.
We combined a model with a concatenated context 
and a model with a single-embedding context, 
or two models with distinct RNN cell types.
A more detailed analysis of the model variants can be found in Appendix \ref{appdx:defmod-analysis}.

\begin{table}
\caption{\label{tab:rd-approaches} Characteristics of our \revdict (RD) approaches (BS = batch size; ME = max epochs; HP = hyperparameter optimization points; S = scheduler; L = loss; MT = multi-task learning).}
\vspace{-2mm}
\small
\centering
\begin{tabular}{l|lccccc}
\hline
\textbf{RD} & \textbf{BS} & \textbf{ME} & \textbf{HP} & \textbf{S} & \textbf{L} & \textbf{MT} \\
\hline
  1 & 1024 & 20  & 30 & CS & MSE & no  \\
  2 & 2048 & 20  & 30 & CS & MSE & no  \\
  3 & 4096 & 20  & 30 & CS & MSE & no  \\
  4 & 8192 & 20  & 30 & CS & MSE & no  \\
  5 & 2048 & 150 & 10 & PS & MSE & no  \\
  6 & 2048 & 150 & 10 & PS & MSE & yes \\
\hline

\end{tabular}
\end{table}

{
\setlength{\tabcolsep}{0.24em}
\begin{table*}[h!]
\caption{Results for the IRB-NLP team systems on the \defmod task. 
MoverScore, BLEU, and lemma-BLEU results are given for each of the five languages.  
Best result across all teams and models is given, followed by the
results of our two best systems. Overall best results of our team are bolded and
the rankings can be found in Table \ref{tab:results}.}
\vspace{-2mm}
\centering
\small
\begin{tabular}{llllllllllllllll}
\toprule
 & \multicolumn{3}{c}{EN} & \multicolumn{3}{c}{ES} & \multicolumn{3}{c}{FR} & \multicolumn{3}{c}{IT} & \multicolumn{3}{c}{RU} \\
 \cmidrule(lr){2-4} \cmidrule(lr){5-7} \cmidrule(lr){8-10} \cmidrule(lr){11-13} \cmidrule(lr){14-16}
  &  \mvsc & \sbleu & \lbleu &  \mvsc & \sbleu & \lbleu &  \mvsc & \sbleu & \lbleu &  \mvsc & \sbleu & \lbleu &  \mvsc & \sbleu & \lbleu \\
\midrule
\besttrow
BEST & 0.135 & 0.033 & 0.043 & 0.128 & 0.045 & 0.064 & 0.075 & 0.029 & 0.038 & 0.117 & 0.066 & 0.099 & 0.148 & 0.049 & 0.072 \\
IRBv3 & 0.089 & 0.032	& 0.040	& 0.093	& \bst{0.045} & \bst{0.064}	& 0.055	& 0.026	& 0.032	& 0.074	& 0.009	& 0.014	& 0.080	& 0.027	& 0.035 \\
IRBv4 & 0.094 & \bst{0.033} & 0.042 &	0.092 &	0.044 & 0.062 & 0.056 & 0.028	& 0.033	& 0.077	& 0.010	& 0.015	& 0.078 & 0.027 & 0.036 \\

\bottomrule
\end{tabular}
\label{table:defmodbest}
\vspace{-4mm}
\end{table*} 
}

\subsection{Reverse Dictionary}
\label{sect:revdict-experimental}

We conducted various development experiments before deciding on the final configuration of our \revdict solutions. 
In all of the experiments, we used the entire set of train data to train the model, and the entire set of validation (development) data for scoring. 
We used \emph{Mean Squared Error} (MSE) as a loss function during training. We tested the effect of cosine loss if added to MSE with different coefficients, but we obtained the best results without cosine loss. We also used MSE for scoring models during 
Bayesian hyperparameter optimization (BHO).

To determine the optimal model size, we searched the space of two transformer hyperparameters: the number of heads and the number of layers. We used a grid search approach with these values $v \in \{1,2,4,8\}$ for both hyperparameters. Additionally, we used BHO \cite{snoek-2012-practical-bayesian-optimization-ml} to find the optimal model for each grid point. However, the increase in model size did not increase the performance of the model. These results were in line with the expectations we had due to the small size of the datasets. Accordingly, we decided to use a transformer with two heads and two layers.
Additionally, we experimented with the maximum length of the input sequence and achieved better validation performance with 256 tokens than 512 with tokens.

We compared performance with and without token embeddings initialization with GloVe vectors. Contrary to our expectations, there was no significant difference in validation performance between these two options, so we skipped the GloVe initialization in the \revdict system settings.
Another development experiment we conducted was to find the optimal method for aggregating the output vectors produced by the transformer, described in Section \ref{sect:system-overview-rd}. We found that the \emph{average} method gives the best results in all cases. 
Furthermore, we examined the influence of the number of layers in the output module on the final prediction. According to the results, there is no benefit in increasing the number of layers in the output module, so we chose a single-layer fully connected network.
We also chose Rectified Linear Unit (ReLU) activation function for the output regression module, because it yielded better performance than hyperbolic tangent (Tanh) activation.

Finally, we made six different solutions for \revdict task. All of these solutions used a two-head transformer architecture, where each head consists of two layers. We used a vocabulary size of 8000 tokens and a maximum sequence length of 256 tokens. The unigram SentencePiece tokenizers used were trained on lowercased but otherwise unmodified glosses contained in a train set. 
We used the \emph{average} method for combining an encoder's output representations and fed them to the output module, which is a single fully-connected layer with RELU activation functions.
We varied five hyperparameters between solutions (Table \ref{tab:rd-approaches}): batch size, max. epochs, number of BHO points, scheduler type, and learning approach.
We used the \emph{Cosine Annealing with Linear Warmup} scheduler (CS) for the first four solutions, and the \emph{Plato} scheduler (PS) for the final two solutions.
We utilized BHO \cite{snoek-2012-practical-bayesian-optimization-ml} to automatically search for optimal hyperparameters
and submitted for testing only models with the best MSE validation scores.

\section{Results}
\label{sect:results}

\newcommand{\mse}{MSE}
\newcommand{\cosm}{COS}
\newcommand{\rnk}{RNK}

{
\setlength{\tabcolsep}{0.20em}
\begin{table*}[h!]
\caption{Results for the IRB-NLP team systems on the \revdict task. 
The best result over all teams and models is given (BEST), followed by the best results of our team (IRB-all) and results of our two specific approaches, IRB-v1 and IRB-v6. 
Finally, the ranks of our team are given (and the number of teams competing in a subtask).
}
\vspace{-2mm}
\centering
\small
\begin{tabular}{lllllllllllllllll}
\toprule
 &  & \multicolumn{3}{c}{EN} & \multicolumn{3}{c}{ES} & \multicolumn{3}{c}{FR} & \multicolumn{3}{c}{IT} & \multicolumn{3}{c}{RU} \\
 \cmidrule(lr){3-5} \cmidrule(lr){6-8} \cmidrule(lr){9-11} \cmidrule(lr){12-14} \cmidrule(lr){15-17}
 &  &  \mse & \cosm & \rnk &  \mse & \cosm & \rnk &  \mse & \cosm & \rnk &  \mse & \cosm & \rnk &  \mse & \cosm & \rnk \\
\midrule
\besttrow
\cellcolor{White}\multirow{4}{*}{\sgns} & BEST & 0.854 & 0.260 & 0.231 & 0.858 & 0.403 & 0.167 & 1.026 & 0.342 & 0.193 & 1.031 & 0.380 & 0.165 & 0.528 & 0.424 & 0.150 \\
 & IRB-all & 0.964 & \bst{0.260} & \bst{0.231} & 0.883 & 0.367 & 0.197 & 1.068 & \bst{0.342} & \bst{0.193} & 1.076 & \bst{0.380} & \bst{0.165} & 0.568 & 0.421 & \bst{0.150} \\
 & IRB-v1 & 1.024 & 0.250 & 0.247 & 0.941 & 0.362 & 0.197 & 1.068 & 0.342 & 0.214 & 1.076 & 0.380 & 0.165 & 0.568 & 0.412 & 0.161 \\
 & IRB-v6 & 1.119 & 0.214 & 0.262 & 1.020 & 0.354 & 0.201 & 1.319 & 0.255 & 0.262 & 1.318 & 0.339 & 0.187 & 0.653 & 0.381 & 0.150 \\
& IRB-rnk & 9 (9) & \bst{1} (9) & \bst{1} (9) & 3 (7) & 2 (7) & 2 (7) & 3 (6) & \bst{1} (6) & \bst{1} (6) & 3 (7) & \bst{1} (7) & \bst{1} (7) & 4 (6) & 2 (6) & \bst{1} (6) \\
\midrule
\besttrow 
\cellcolor{White}\multirow{4}{*}{\chr} & BEST &  0.141 & 0.798 & 0.419 & 0.467 & 0.839 & 0.403 & 0.335 & 0.789 & 0.416 & 0.334 & 0.747 & 0.383 & 0.116 & 0.852 & 0.357 \\
 & IRB-all &  0.162 & 0.770 & \bst{0.419} & 0.526 & 0.819 & \bst{0.403} & 0.390 & 0.756 & 0.421 & 0.366 & 0.724 & \bst{0.383} & 0.140 & 0.824 & \bst{0.357} \\
 & IRB-v1 & 0.169 & 0.761 & 0.438 & 0.526 & 0.819 & 0.407 & 0.409 & 0.744 & 0.425 & 0.366 & 0.724 & 0.397 & 0.145 & 0.818 & 0.361 \\
 & IRB-v6 &  0.172 & 0.765 & 0.444 & 0.635 & 0.784 & 0.420 & 0.434 & 0.734 & 0.421 & 0.399 & 0.711 & \bst{0.383} & 0.144 & 0.821 & \bst{0.357} \\
 & IRB-rnk & 5 (7) & 7 (7) & \bst{1} (7) & 3 (5) & 5 (5) & \bst{1} (5) & 3 (5) & 4 (5) & 2 (5) & 5 (6) & 5 (6) & \bst{1} (6) & 3 (5) & 4 (5) & \bst{1} (5) \\
\midrule
\besttrow 
\cellcolor{White}\multirow{4}{*}{\electra} & BEST &  1.301 & 0.847 & 0.432 &  & &  & 1.066 & 0.862 & 0.429 &  & &  & 0.828 & 0.735 & \bst{0.345}
 \\
 & IRB-all &  1.685 & 0.828 & \bst{0.432} & & & & 1.339 & 0.847 & \bst{0.429} & & & & 0.911 & 0.724 & \bst{0.345} \\
 & IRB-v1 &  1.723 & 0.821 & 0.438 &  & &  & 1.339 & 0.847 & 0.447 &  & &  & 0.911 & 0.724 & 0.350
 \\
 & IRB-v6 &  1.988 & 0.792 & \bst{0.432} &  & &  & 1.566 & 0.825 & \bst{0.429} &  & &  & 1.049 & 0.702 & \bst{0.345} \\
 & IRB-rnk & 6 (6) & 6 (6) & \bst{1} (6) &  & & &  4 (4) & 4 (4) & \bst{1} (4) &   & & &  4 (4) & 3 (4) & \bst{1} (4) \\
\bottomrule
\end{tabular}
\label{table:revdictbest}
\vspace{-3mm}
\end{table*} 
}

\paragraph{Definition Modeling}

On the \defmod task, the models were evaluated using three metrics: 
BLEU score \cite{papineni-2002-bleu}, lemma-level BLEU score, 
and MoverScore \cite{zhao-2019-moverscore}. While the BLEU score is based on 
matching token n-grams between the reference and the model-produced text, 
MoverScore calculates a measure of distance
between texts embedded in a semantic space, i.e., between two sets of contextual word embeddings
computed using a transformer model.

Table \ref{table:defmodbest} contains scores for two of our best model configurations, 
``version 3'' and ``version 4''.
Both model configurations are described in detail at the
end of Section \ref{sect:defmod-experim}.
While version 3 models are based on GRU RNN and trained using $300$ training epochs, version 4 models are built with either GRU or LSTM and $450$ epochs. 
The fallback strategy, which yields slight performance gains, is also used.
These results are presented and analyzed in a 
more detailed manner in Appendix \ref{appdx:defmod-analysis}.
Results in Table \ref{table:defmodbest} show that our models are competitive 
with other teams' models on English, Spanish and French, 
especially in terms of the BLEU scores. 
MoverScore results are weaker than those produced by the top models, 
but rank among the upper half of the systems 
except for Italian and Russian, languages for which our models' performance is below average.
Rankings aggregated across all the scores, displayed in Table \ref{tab:results}, reflect the above observations and show that the models we produced can perform quite competitively.

Our approach shows inter-language variation, both in relative (ranks)
and absolute (score values) terms. 
The full results provided by the organizers\footnote{\codwoerepo}
show that this is also true for other teams -- for example, few of the high-performing models perform 
markedly better for Italian and Russian than for other languages.
However, some approaches yield more stable results across all languages.

All of the models yielded by the \codwoe shared task perform weakly in terms of BLEU.
Namely, the BLEU scores of the existing \defmod approaches commonly achieve
BLEU scores in the range of $20$ to $30$ \cite{noraset-2017-dm-definition, kabiri-2020-dm-evaluating}, with some settings yielding BLEU as high as $60$ \cite{kabiri-2020-dm-evaluating}. 
The experiments with the weakest reported BLEU scores \cite{gadetsky-2018-dm-conditional, kabiri-2020-dm-evaluating} reports BLEU scores of approx. $12$, 
while the best \codwoe scores are below BLEU $10$.

\codwoe \defmod models perform better in terms of MoverScore, 
a metric designed for machine summarization \cite{zhao-2019-moverscore}.
An analysis of a number of summarization systems showed that MoverScore values range between $15$ and $24$, with an absolute minimum of $10$ and an average slightly below $20$ \cite{fabbri2021-sumeval}.
In comparison, top \codwoe systems reach scores between $12$ and $15$, except in the case of French, 
which puts them on the lower end of the summarization scale.

We hypothesize that the main reason for the described weak performance is 
comparatively small amount of \codwoe training data (for each individual language), 
as well as the lack of word embeddings pre-trained on a large outside corpus.
Namely, most of the other \defmod approaches use at least $2$--$3$ times more training data, 
both in terms of the number of (embedding, text) examples, and the overall number of tokens
\cite{noraset-2017-dm-definition, gadetsky-2018-dm-conditional, zhu_multi-sense_2019, mickus_mark_2019, bevilacqua-2020-dm-generationary, zhang-2020-dm-interpretability, yang-2020-dm-sememes}.
Additionally, these approaches make use of the pre-trained word embeddings 
that carry the semantic information extracted from a huge corpus.

As for the representativeness of the test data, the visual analysis performed in \ref{appdx:defmod-data}
shows that the distribution of test gloss embeddings matches the train distribution well.
Another factor that potentially influences performance is word rarity.
We observed that the English test examples contain a significant amount of rare words (such as ``pelta'', ``akimbo'', ``gothy'', or ``dungarees''), 
while some \defmod experiments explicitly focus on the most frequent words \cite{noraset-2017-dm-definition}.

The greatest performance gains for the models we used come from
using all three vector embeddings to form a context vector. 
This suggests that future approaches can benefit from leveraging 
several distinct embeddings types as input for gloss generation.

We believe that the question of the influence of various factors on the 
performance of \defmod systems is important and under-explored.
These factors include model structure and parameters,  performance metric, 
dataset size (both for training and pre-training), and the semantic relation between training and test data.
Closely related is the question of the nature of semantic generalization that \defmod systems are capable of --
what kind of examples (and relations contained within them) can inform 
a successful inference of glosses for unseen embeddings.

Further performance-related analyses can be found in Appendix \ref{appdx:defmod-analysis}. Appendix \ref{app:gloss-analysis} contains a qualitative analysis of glosses that shows that generated glosses can capture varying levels of semantic properties of the correct glosses. We hypothesize that these variations in similarity are hard to capture with metrics such as MoverScore and BLEU.

\paragraph{Reverse Dictionary}
We used the following metrics for internal validation of our \revdict solutions (described in Section \ref{sect:revdict-experimental}): \emph{Mean Squared Error} (MSE), \emph{Cosine Similarity} (COS), and \emph{Central Kernel Alignment} (CKA) \cite{kornblith-2019-cka, cortes-2012-cka}. COS measure has noted drawbacks \cite{heidarian-2016-similarity}. Therefore, we use the linear CKA similarity measure to gain another perspective on model performance. Validation scores
can be found in Appendix \ref{appdx:revdict-model-analysis}, Table \ref{tab:rd-val-score-all}.
It is evident that each subsequent approach gives better validation results than the previous ones. 

Test predictions were scored by the following metrics: MSE, COS, and \emph{Cosine-Based Ranking} (RNK). The RNK measure is defined as the proportion of test samples with cosine similarity to the model output embedding higher than the ground truth embedding. The final results of our solutions can be found in Table \ref{tab:rd-test-score-all} (see Appendix \ref{appdx:revdict-model-analysis}). Here, each subsequent approach has lower scores than the previous ones, which is the complete opposite of the validation results. This suggests potential overfitting to the dev dataset that could be the result of BHO. However, this is contrary to expectations as the last two solutions have three times fewer BHO points and should not overfit to the dev dataset. The reason for this phenomenon is unclear and needs further investigation.
Finally, the best \revdict results for each team can be found in Appendix \ref{appdx:revdict-model-analysis} (Table \ref{tab:rd-test-mse-all-teams} for MSE, Table \ref{tab:rd-test-cos-all-teams} for COS, and Table \ref{tab:rd-test-rnk-all-teams} for RNK).
The test results and overall rankings of our solutions are summarized in Table \ref{table:revdictbest}.

\begin{figure}
    \caption{Example of two different predictions for ground truth vector $V_{GT}$, where predicted vector $V_1$ has better MSE and COS scores than $V_2$, and $V_2$ has better RNK score than $V_1$. The rest of the points represent vectors of other test samples.}
\vspace{-2mm}
    \label{fig:rd-rnk-example}
    \centering
    \includegraphics{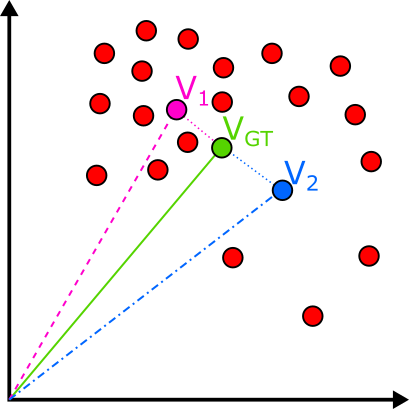}
\vspace{-6mm}
\end{figure}

Compared to other solutions, our systems have average or below-average performance in terms of MSE and COS test scores. However, they perform significantly better than the other approaches in terms of RNK test scores, from which we conclude that our solutions are better suited for the retrieval task. This is an interesting situation which we elaborate with the following example, shown in Figure \ref{fig:rd-rnk-example}. It depicts two different predictions, $V_1$ and $V_2$, the first with better MSE and COS scores, and the second with a better RNK score. The second solution prefers a vector subspace with a lower density of test samples even if the absolute distance from the correct vector is greater. With a smaller set of possible surrounding solutions, retrieving the vector $V_{GT}$ from the vector $V_2$ is more precise than retrieving it from the vector $V_1$.

\section{Conclusion}
Definition Modeling and Reverse Dictionary are two opposite learning tasks for exploring the relationship between different semantic representations of words.
\codwoe SemEval task \cite{codwoe-paper} is designed to investigate
these tasks on five different languages using three different types of word embeddings.

We propose an adaptation of an existing \defmod model and analyze its 
performance and the glosses generated by the model.
We believe that \defmod is a theoretically interesting problem 
and that further investigations should focus on discovering
which types of semantic generalization the models are able to perform, 
and how this generalization ability is influenced by both the data 
and the models' structure. The existing \defmod experiments are largely 
incomparable since they are based on different data and setups.
We believe that a contribution of the \codwoe task is the
creation of a multilingual evaluation setting, 
as well as the use of the flexible MoverScore as an evaluation metric.

Our \revdict systems are based on deep regression models based on transformer architecture that
achieved top scores for the difficult-to-predict \sgns (word2vec)  embeddings.
In most cases our \revdict solutions perform significantly better then the other systems in terms of the RNK score.
These results imply that our solutions could be the appropriate approach for retrieving the right word from its description, a problem crucial for solving the TOT problem \cite{brown-1966-tip-of-the-tongue} in machine-assisted text synthesis.

In summary, the models that we produced for the \codwoe task 
perform competitively when compared to other participants'
models, and can therefore serve as a reasonable 
starting point for future tackling of \defmod and \revdict problems.
We believe that the promising directions for future optimizations include the
construction of multilingual and multi-task models, as well as investigations
of the influence of the external data, primarily in the form of huge pre-training corpora.

\section*{Acknowledgments}
We would like to thank the anonymous reviewers for the useful comments, 
as well as Timothee Mickus for the help with challenge-related questions.
We would also like to thank Miha Keber and Tomislav Lipić for helpful 
discussions and advice.

\newpage

\bibliographystyle{acl_natbib}
\bibliography{custom}

\appendix

\clearpage

\onecolumn
\section{Appendix - Analysis of \defmod Data and Models}

\subsection{Train and Test Data}
\label{appdx:defmod-data}

Motivated by the weak performance of \defmod models
(see Section \ref{sect:results}), we 
examined whether the distributions of train and test data 
are comparable. To this end we created 2D projections of \sgns and \electra
embedding for all five languages using the t-SNE method \cite{maaten2008tsne}.

The projections, depicted in Figure \ref{fig:emb-projections}, 
show that the train and test distributions of the embeddings match well. 
It is therefore reasonable to expect that the distributions of the 
gloss texts are similar as well, as the gloss semantics 
expectedly matches the semantics of the corresponding words.
However, this conjecture should be confirmed experimentally, 
for example by per-gloss aggregation of pretrained word embeddings
extracted from huge corpora.

Figure \label{fig:emb-projections} also shows that the \electra
vectors are more separable than the \sgns vectors. 
The separability of the embedding vectors varies across languages, 
probably influenced by the corpora used for pre-training of the embeddings.
We note that the observations about the train and test embedding
distributions are also applicable to the \revdict problem 
aimed at the prediction of the embeddings from gloss texts.

Basic gloss statistics can be found in Table \ref{tab:orig-gloss-stats}. 
There exists a large variation in gloss size between languages, e.g., 
the longest gloss from the ES dataset is almost twice 
the size of the longest EN gloss. 
In addition, the longest glosses in the validation (development) datasets are significantly smaller then those in the train datasets, on average $42.55\%$ smaller.
The 'dictionary size' column in the table is the number of distinct tokens in each dataset. Dictionary sizes vary, for example, EN dictionary is approximately half the size of the RU dictionary.
Differences between the gloss and dictionary sizes suggest 
that it is reasonable to use a separate model for each language.

Basic statistics of the transformed dataset can be found in 
Table \ref{tab:modified-gloss-stats}. As expected,  
the transformed glosses are significantly smaller 
then the glosses in the original dataset. For example, the median 
transformed gloss size is on average $29.25\%$ smaller.

\newcommand{\figw}{65mm}
\newcommand{\rows}{6pt}

\begin{figure*}[h!]
\caption{t-SNE projections of the \sgns and \electra vectors from the 
train (green) and test (red) datasets. Color intensity is 
proportional to data density.}
\begin{tabular}{cc}
  \includegraphics[width=\figw]{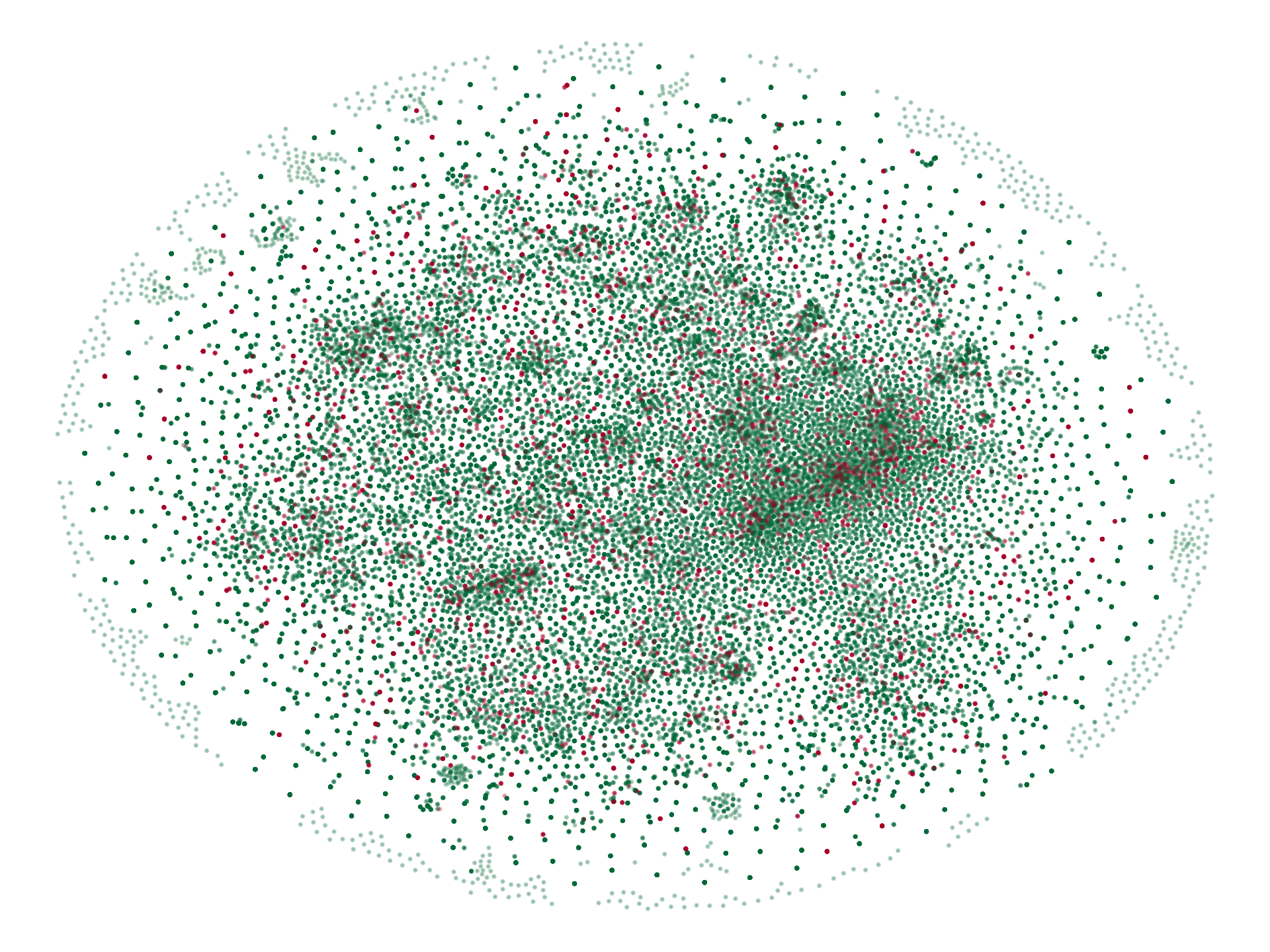} &   \includegraphics[width=\figw]{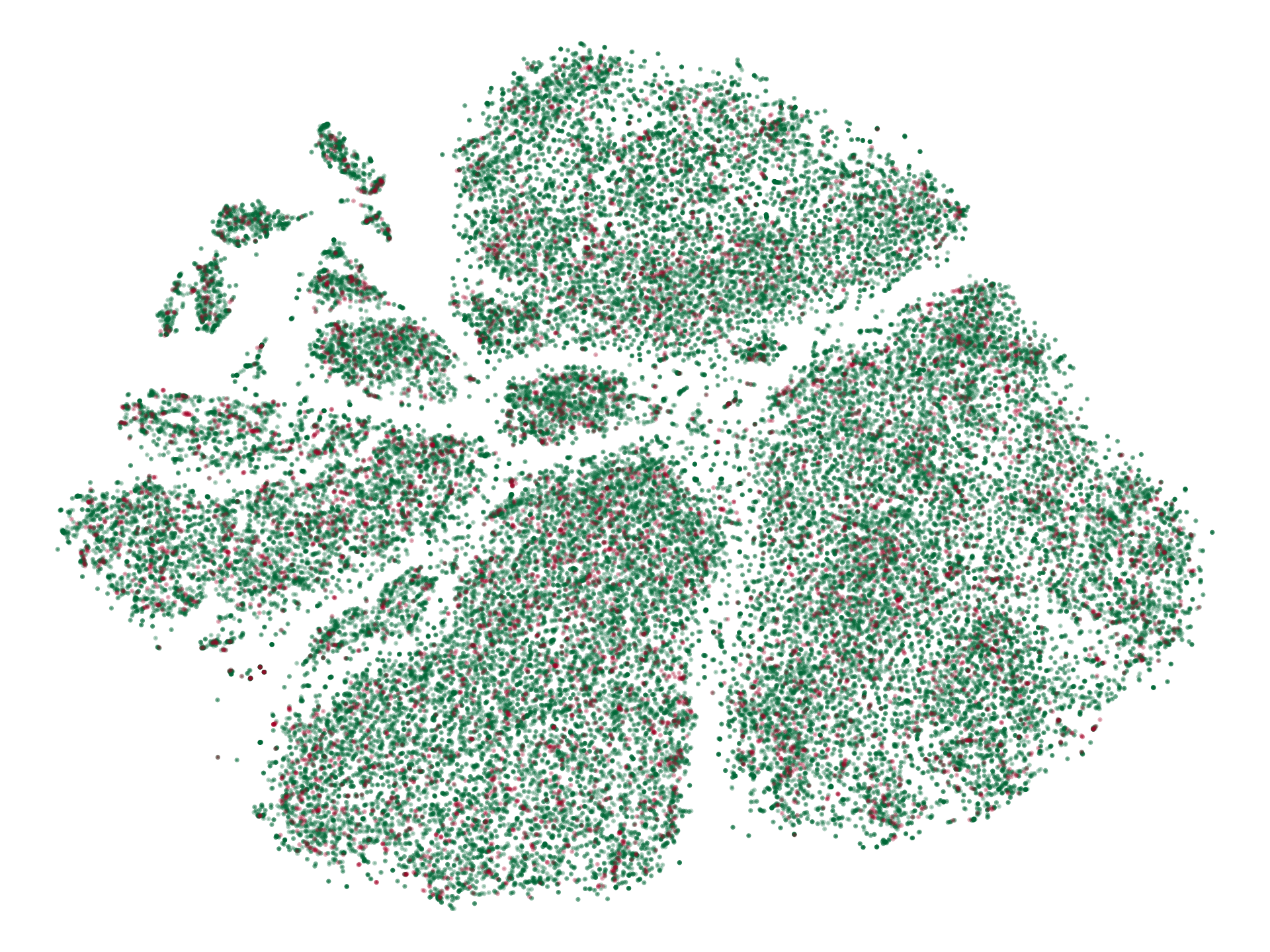} \\
EN-SGNS & EN-ELECTRA \\[\rows]
 \includegraphics[width=\figw]{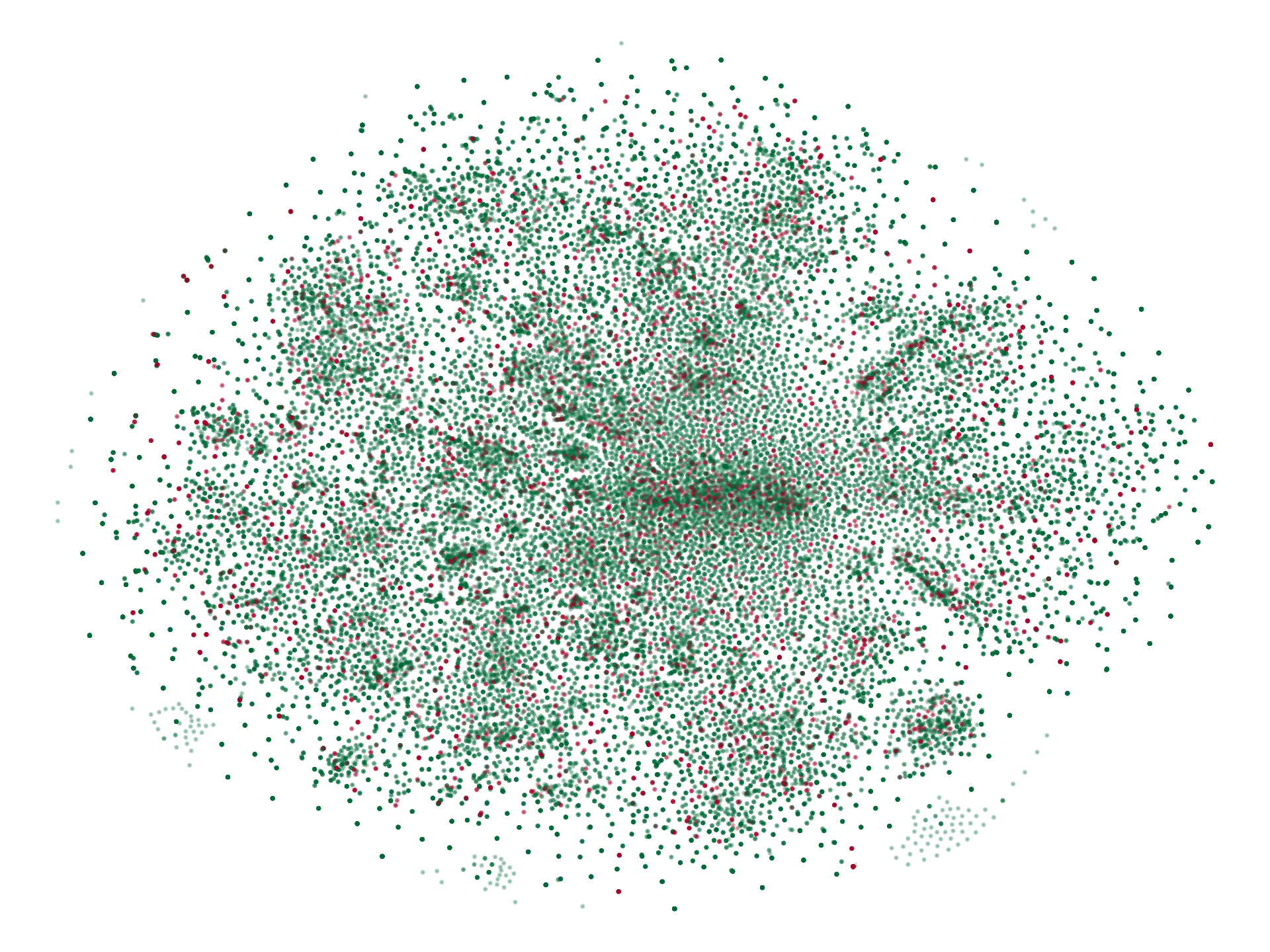} &   \includegraphics[width=\figw]{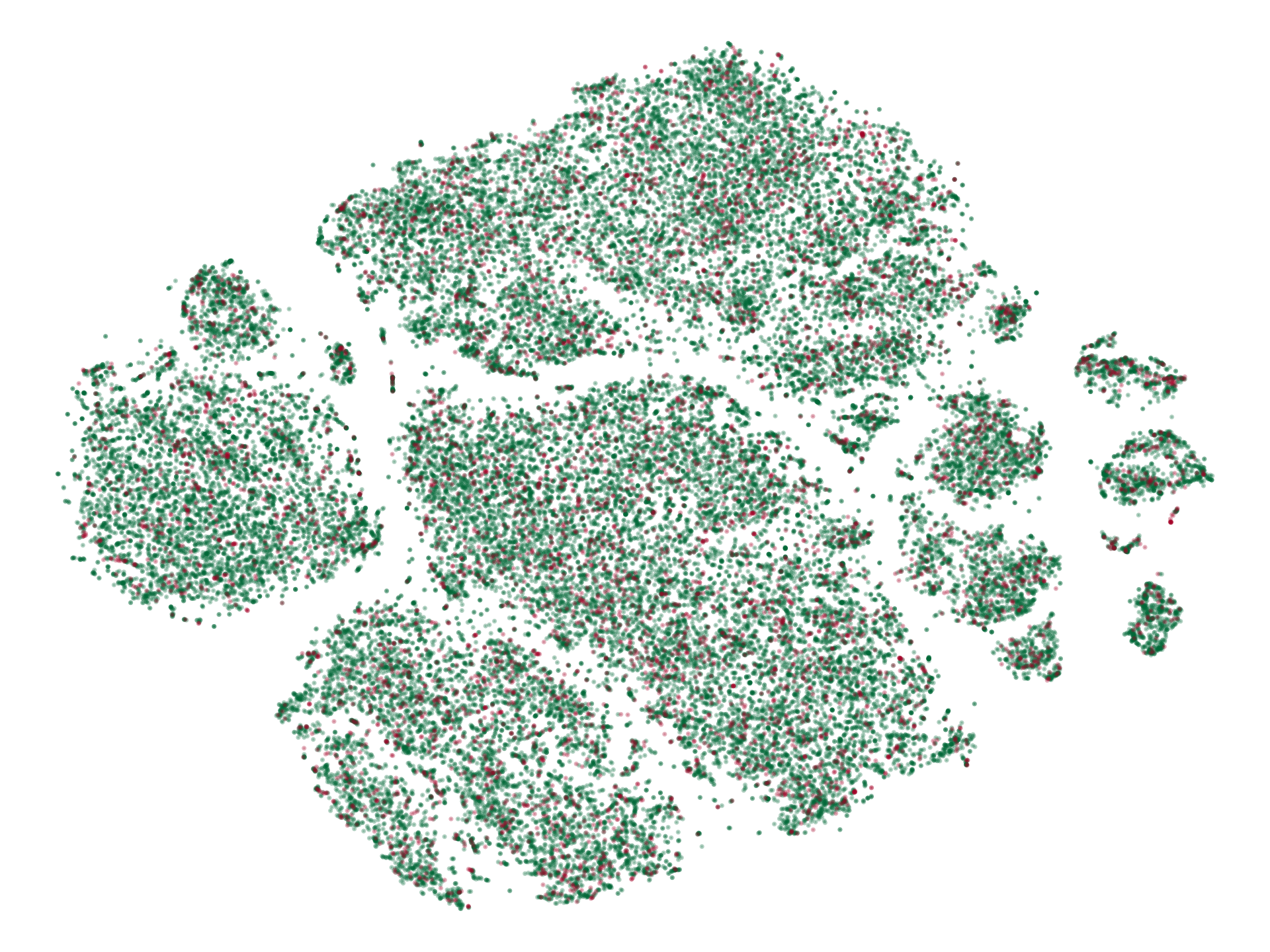} \\
FR-SGNS & FR-ELECTRA \\[\rows]
 \includegraphics[width=\figw]{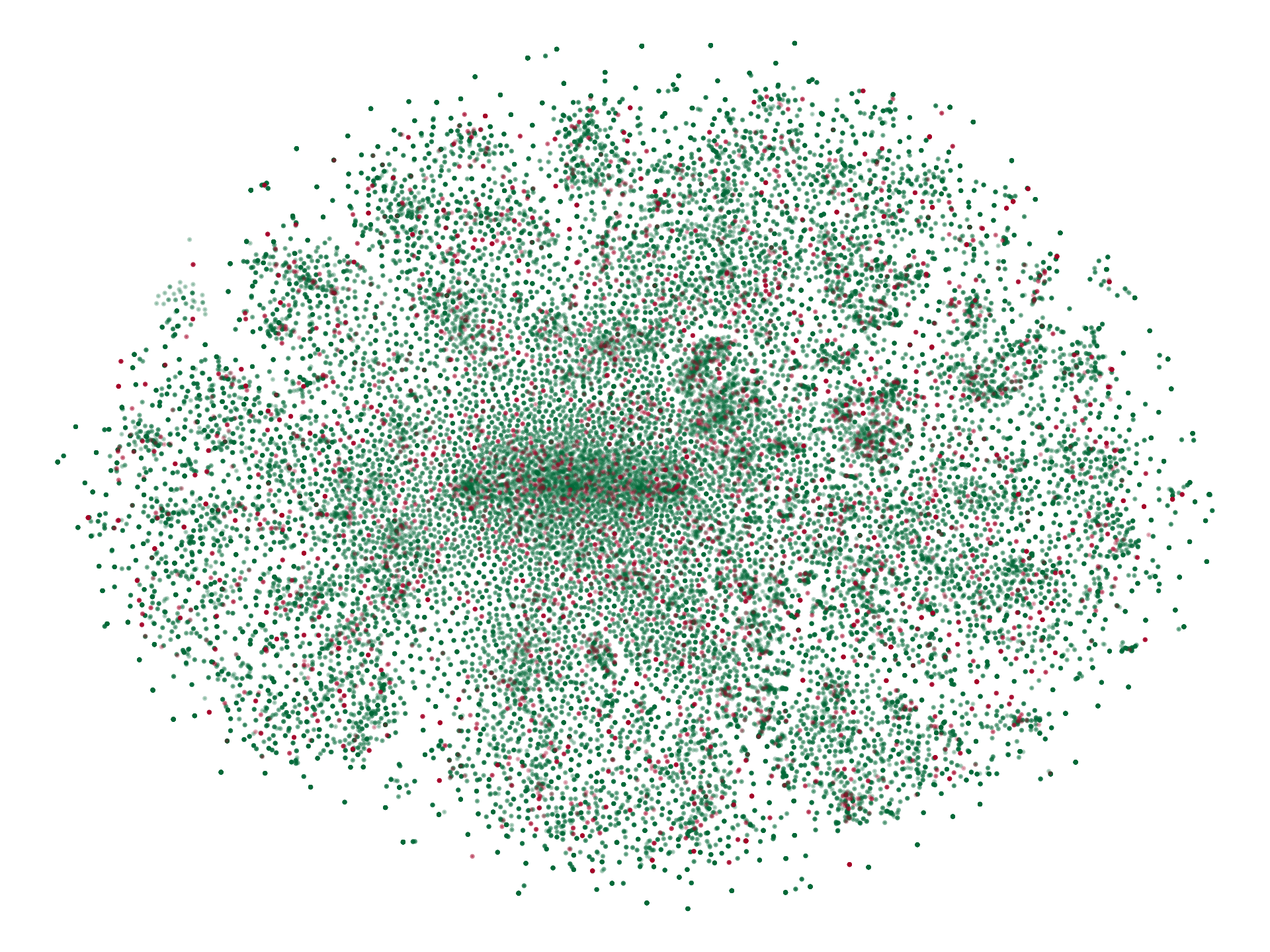} &   \includegraphics[width=\figw]{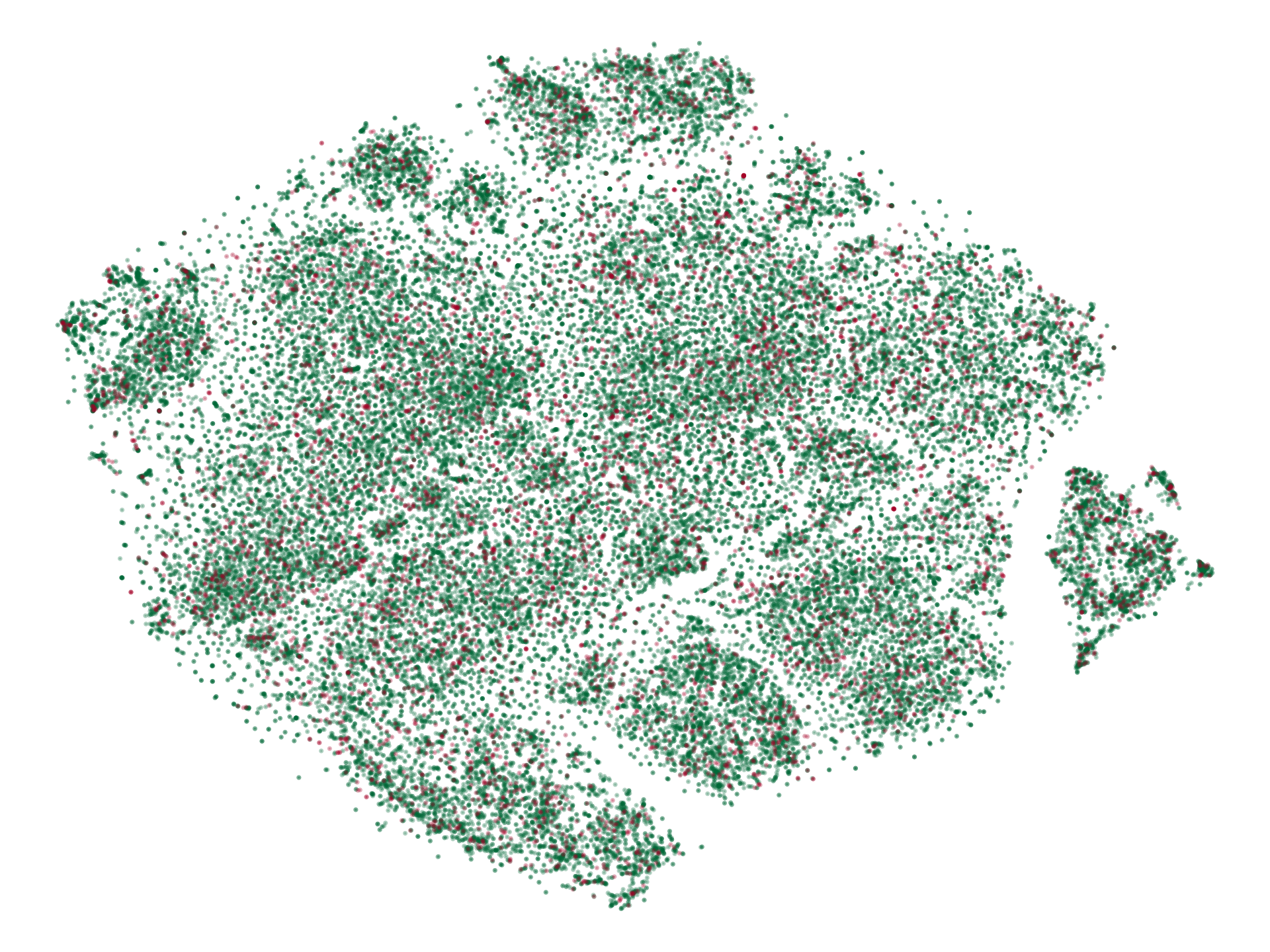} \\
RU-SGNS & RU-ELECTRA \\[\rows]
 \includegraphics[width=\figw]{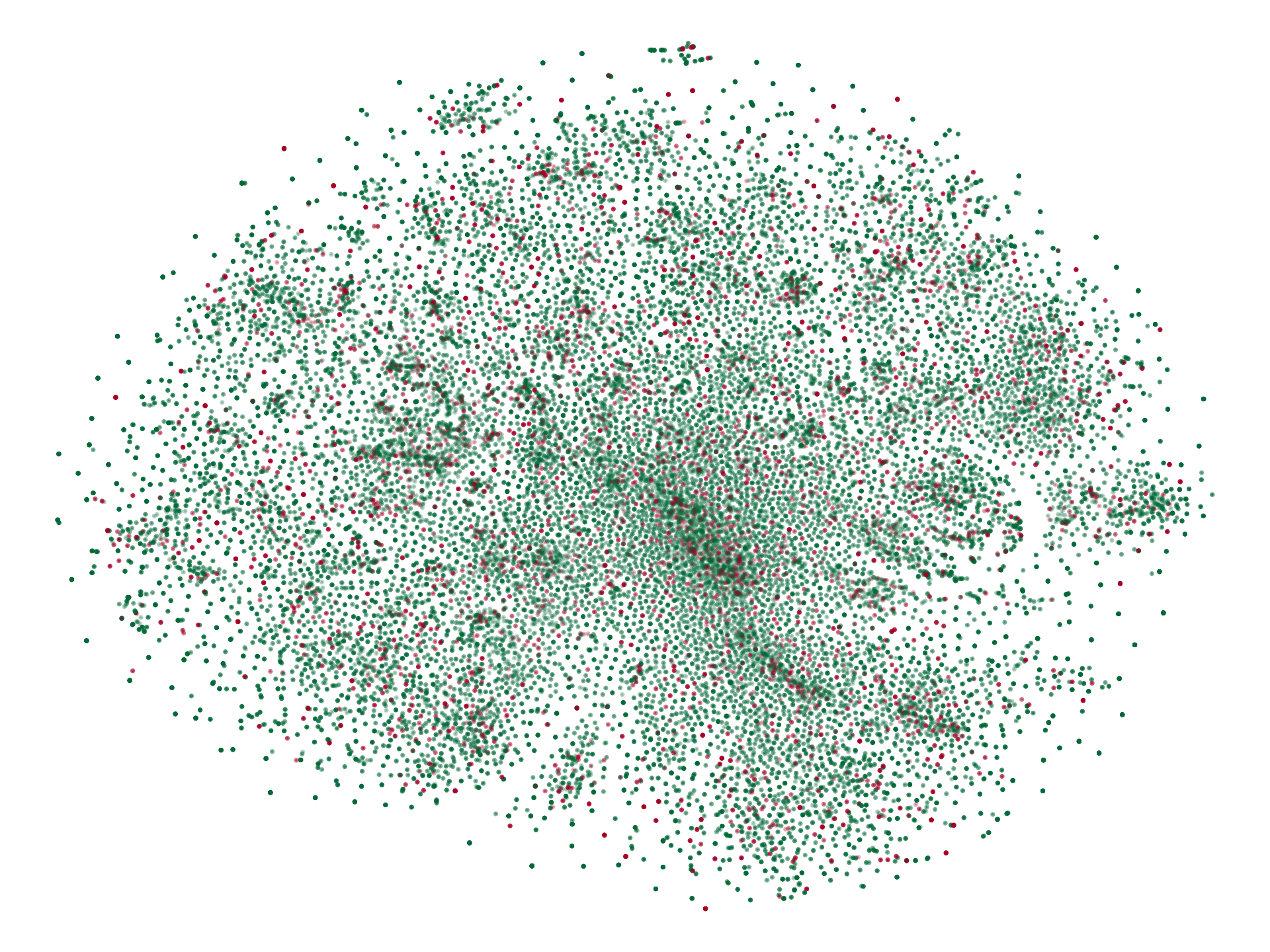} &   \includegraphics[width=\figw]{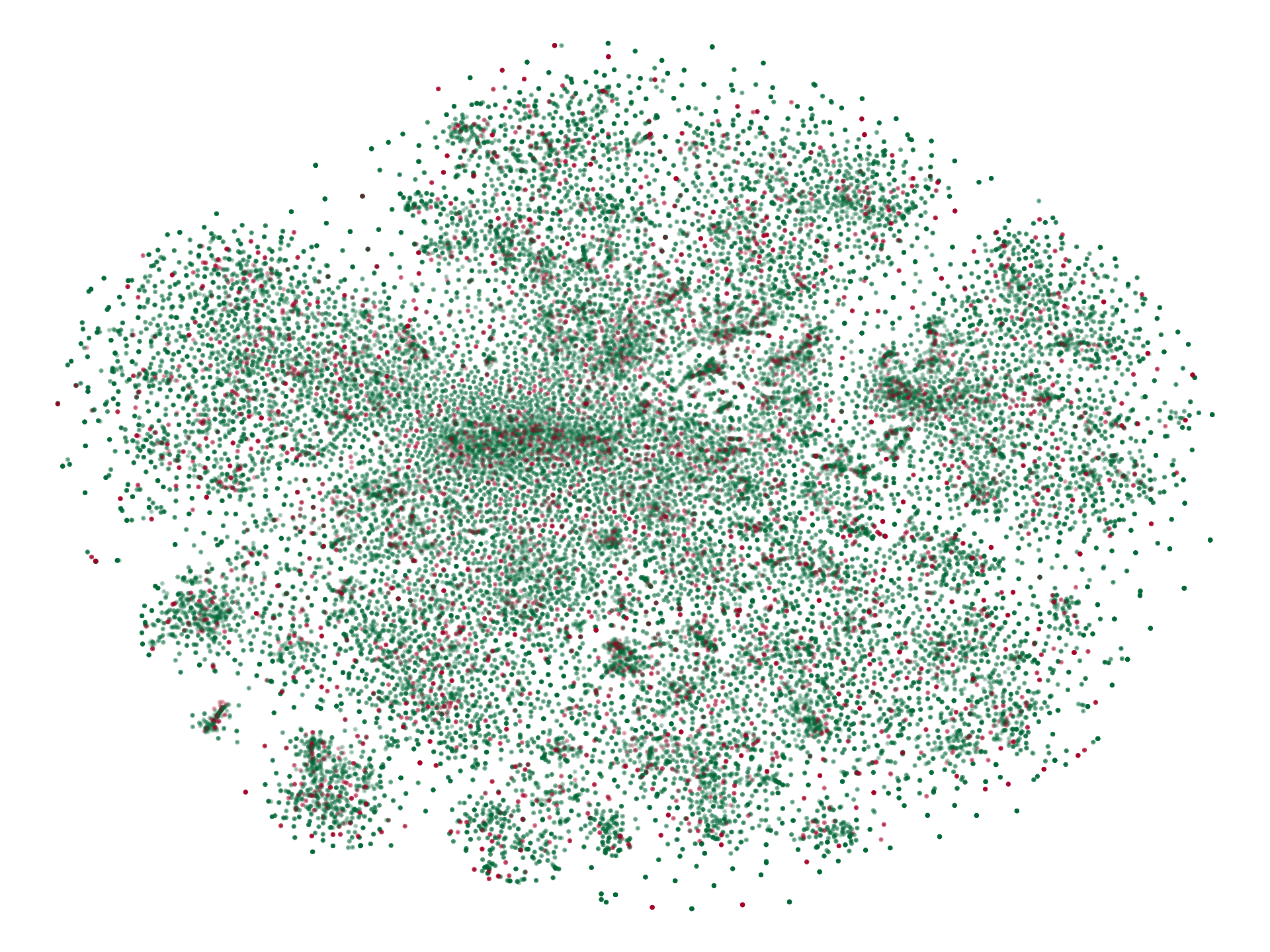} \\
ES-SGNS & IT-SGNS \\[\rows]
\end{tabular}
\label{fig:emb-projections}
\end{figure*}

\begin{table*}[h!]
\centering
\begin{tabular}{llrrrrrrrrrr}
\hline
Lang. & Split & Dict. size & \#Tokens & \#Glosses & \multicolumn{7}{c}{Gloss size} \\
 &  &  &  &  & mean & st.dev & min & q25 & median & q75 & max \\
\hline
EN & train & 29.046 & 511.531 & 43.608 & 11.73 & 7.98  & 1 & 6.0 & 10.0 & 15.0 & 129 \\
EN & dev   & 9.478  & 76.073  & 6.375  & 11.93 & 7.98  & 1 & 6.0 & 10.0 & 15.0 & 70  \\
ES & train & 46.765 & 647.093 & 43.608 & 14.84 & 13.07 & 1 & 7.0 & 11.0 & 18.0 & 257 \\
ES & dev   & 15.464 & 91.943  & 6.375  & 14.42 & 12.22 & 1 & 7.0 & 11.0 & 17.0 & 159 \\
FR & train & 40.032 & 623.978 & 43.608 & 14.31 & 9.74  & 1 & 8.0 & 12.0 & 18.0 & 159 \\
FR & dev   & 12.760 & 91.475  & 6.375  & 14.35 & 9.91  & 1 & 8.0 & 12.0 & 18.0 & 113 \\
IT & train & 40.130 & 592.409 & 43.608 & 13.58 & 11.01 & 1 & 6.0 & 11.0 & 18.0 & 202 \\
IT & dev   & 14.069 & 87.531  & 6.375  & 13.73 & 11.61 & 1 & 6.0 & 11.0 & 18.0 & 130 \\
RU & train & 57.141 & 492.978 & 43.608 & 11.30 & 7.78  & 1 & 6.0 & 9.0  & 14.0 & 169 \\
RU & dev   & 15.498 & 70.392  & 6.375  & 11.04 & 7.22  & 1 & 6.0 & 9.0  & 14.0 & 74  \\
\hline
\end{tabular}
\caption{Statistics of the gloss and dictionary sizes for the original train and validation (development) datasets. Sizes are calculated by counting the number of whitespace-delimited tokens.}
\label{tab:orig-gloss-stats}
\end{table*}

\begin{table*}[h!]
\centering
\begin{tabular}{llrrrrrrrrrr}
\hline
Lang. & Split & Dict. size & \#Tokens & \#Glosses & \multicolumn{7}{c}{Gloss size} \\
 &  &  &  &  & mean & st.dev & min & q25 & median & q75 & max \\
\hline
EN & train & 25.921 & 456.673 & 58.792 & 7.77  & 6.97  & 1 & 3.0 & 6.0  & 10.0 & 128 \\
EN & dev   & 8.892  & 68.145  & 8.403  & 8.11  & 7.06  & 1 & 3.0 & 6.0  & 11.0 & 69  \\
ES & train & 40.024 & 595.879 & 44.543 & 13.38 & 12.01 & 1 & 6.0 & 10.0 & 16.0 & 168 \\
ES & dev   & 13.723 & 84.303  & 6.493  & 12.98 & 11.32 & 1 & 6.0 & 10.0 & 16.0 & 158 \\
FR & train & 33.963 & 487.013 & 46.537 & 10.47 & 9.18  & 1 & 4.0 & 8.0  & 14.0 & 155 \\
FR & dev   & 11.216 & 71.021  & 6.786  & 10.47 & 9.22  & 1 & 4.0 & 8.0  & 14.0 & 101 \\
IT & train & 39.124 & 452.028 & 45.080 & 10.03 & 9.03  & 1 & 4.0 & 7.0  & 13.0 & 195 \\
IT & dev   & 13.805 & 67.211  & 6.621  & 10.15 & 9.40  & 1 & 4.0 & 7.0  & 13.0 & 109 \\
RU & train & 56.467 & 428.787 & 50.843 & 8.43  & 6.99  & 1 & 4.0 & 7.0  & 11.0 & 142 \\
RU & dev   & 15.241 & 61.074  & 7.509  & 8.13  & 6.44  & 1 & 4.0 & 6.0  & 11.0 & 72  \\
\hline
\end{tabular}
\caption{Statistics of the gloss and dictionary sizes for the transformed train and validation (development) datasets. Sizes are calculated by counting the number of whitespace-delimited tokens.}
\label{tab:modified-gloss-stats}
\end{table*}

\subsection{\defmod Models' Performance}
\label{appdx:defmod-analysis}

Here we append Section \ref{sect:results} with a more fine-grained
analysis of the \defmod models. Table \ref{table:defmoddetails} contains the models' performances.
As can be seen, the largest gains are achieved by using all of the embedding vectors as input for gloss generation (context=\allvec). There exists a negligible difference between the LSTM and GRU RNNs, with GRU performing slightly better.
Using a fallback model always slightly improves the MoverScore of a model.
In Table \ref{table:defmoddetails} the architecture of the fallback model is
the architecture of the main model with the corresponding parameter 
replaced with the value in the 'fallback' column.
Interestingly, using contextual \electra vectors does not help, 
i.e., the \sgns (word2vec) vectors which are not context-aware perform comparably. 
This is true even when only a single embedding is used, i.e., when context equals \electra.
The equality of \sgns and \electra is unexpected since both the train and test datasets contain polysemous \electra vectors and words with multiple senses.

It is also interesting to consider the influence of the 
training data on the model's performance.
We hypothesize that a \defmod model's score on a single test example is positively correlated with the semantic closeness of the example to the examples in the train set. To test this hypothesis we calculate Spearman correlation between test MoverScore and BLEU on one, and the cosine similarity of the test embedding and most similar train embeddings.
This is done for the best-performing submitted model from Table \ref{table:defmoddetails}.
We also calculate the average scores on two sets of $10\%$ test examples
that are least similar and most similar to the train examples.
Since the embeddings (\sgns and \electra) were built on large outside corpora, 
it is reasonable to believe that they capture semantic similarity of the
associated words and glosses.
Surprisingly, the results show a lack of consistent and strong correlation 
and the correlations range from weakly negative to weakly positive, depending on both the language and the embedding type.
This lack of correlation could be caused by many factors, including the nature of the model, the nature of the pretrained embeddings, and the semantics of the cosine similarity measure. 

The future extensions and improvements of the proposed analysis 
could reveal the nature of the train data necessary for the \defmod models to successfully generalize, and perhaps point to a similarity measure that reveals 
more fine-grained properties of such a generalization.

{
\setlength{\tabcolsep}{0.24em}
\begin{table*}[h!]
\caption{Correlation between the best \defmod model's scores on one, 
and the closeness of the test examples to the train set on the other side.
The unit of correlation is an example from the test set, 
and its similarity to the train set is calculated as the average
cosine similarity with the $10$ most similar train embeddings. 
The last two columns contain average model scores on $10\%$ of the least and most train-similar test examples.}
\centering
\normalsize
\begin{tabular}{rrrrrcccc}
\toprule
 & \multicolumn{4}{c}{Correlation of Score and Similarity} & \multicolumn{4}{c}{Avg. Score for Similarity Percentile} \\
 \cmidrule(lr){2-5} \cmidrule(lr){6-9}
 & \multicolumn{2}{c}{\mvsc} & \multicolumn{2}{c}{\sbleu} & \multicolumn{2}{c}{\mvsc} & \multicolumn{2}{c}{\sbleu} \\ 
 \cmidrule(lr){2-3} \cmidrule(lr){4-5} \cmidrule(lr){6-7} \cmidrule(lr){8-9}
LANG-EMB & spearman $\rho$ & p-value & spearman $\rho$ & p-value & bottom 10\% & top 10\% & bottom 10\% & top 10\% \\
 \midrule
EN-SGNS & 0.0458 & 0.0003 & 0.0019 & 0.8831 & 0.0852 & 0.1004 & 0.0328 & 0.0297  \\
EN-ELKT & 0.0096 & 0.4508 & 0.0186 & 0.1414 & 0.0889 & 0.1182 & 0.0293 & 0.0503 \\ 
FR-SGNS & -0.0625 & 0.0000 & -0.1270 & 0.0000 & 0.0801 & 0.0427 & 0.0363 & 0.0222 \\ 
FR-ELKT & 0.0433 & 0.0006 & 0.0838 & 0.0000 & 0.0387 & 0.0760 & 0.0214 & 0.0322 \\ 
RU-SGNS & 0.0758 & 0.0000 & 0.0353 & 0.0054 & 0.0754 & 0.0947 & 0.0310 & 0.0279 \\ 
RU-ELKT & 0.0000 & 0.9979 & -0.0063 & 0.6217 & 0.0748 & 0.0677 & 0.0279 & 0.0244 \\ 
ES-SGNS & 0.0458 & 0.0003 & 0.0019 & 0.8831 & 0.1084 & 0.1052 & 0.0523 & 0.0552 \\ 
IT-SGNS & -0.0528 & 0.0000 & 0.0047 & 0.7084 & 0.1082 & 0.0783 & 0.0111 & 0.0128 \\ 
\bottomrule
\end{tabular}
\label{table:defmodcorr}
\end{table*} 
}

\renewcommand{\bst}[1]{\textcolor{red}{\textbf{#1}}}
\newcommand{\bsts}[1]{\textbf{#1}}
{
\begin{landscape}
\setlength{\tabcolsep}{0.28em}
\begin{table}[h!]
\caption{Performance of best \defmod models. Overall best models from all the participants are included for comparison. Submitted models are colored gray and the best submitted results are bolded. The best overall results are colored red. Other models are included to illustrate the influence of design choices.}
\centering
\normalsize
\begin{tabular}{lllccllllllllllllllll}
\toprule
\multicolumn{5}{c}{MODEL PARAMS} & \multicolumn{3}{c}{EN} & \multicolumn{3}{c}{ES} & \multicolumn{3}{c}{FR} & \multicolumn{3}{c}{IT} & \multicolumn{3}{c}{RU} \\
\cmidrule(lr){1-5} \cmidrule(lr){6-8} \cmidrule(lr){9-11} \cmidrule(lr){12-14} \cmidrule(lr){15-17} \cmidrule(lr){18-20}
 context & seed & rnn & fallback & \#epochs & \mvsc & \sbleu & \lbleu &  \mvsc & \sbleu & \lbleu &  \mvsc & \sbleu & \lbleu &  \mvsc & \sbleu & \lbleu &  \mvsc & \sbleu & \lbleu \\
\midrule
\besttrow
\multicolumn{5}{c}{BEST} & 0.135 & 0.033 & 0.043 & 0.128 & 0.045 & 0.064 & 0.075 & 0.029 & 0.038 & 0.117 & 0.066 & 0.099 & 0.148 & 0.049 & 0.072 \\
\submitrow
\sgns & \sgns & \gru & \none & $300$ & 0.070 & 0.027 & 0.034 & 0.083 & 0.039 & 0.058 & 0.039 & 0.024 & 0.028 & 0.073 & 0.009 & 0.014 & 0.073 & 0.023 & 0.031 \\
\submitrow
\allvec & \sgns & \gru & \none & $300$ & 0.085 & 0.031 & 0.040 & 0.092 & 0.045 & 0.064 & 0.048 & 0.026 & 0.032 & 0.072 & 0.009 & 0.014 & 0.077 & 0.027 & 0.035 \\
\submitrow
\allvec & \sgns & \gru & \sgns & $300$ & 0.089 & 0.032 & 0.040 & \bsts{0.093} & \bsts{0.045} & \bsts{0.064} & 0.055 & 0.026 & 0.031 & 0.074 & 0.009 & 0.014 & \bsts{0.080} & 0.027 & 0.035 \\
\midrule
\sgns & \sgns & \gru & \none & $450$ & 0.077 & 0.029 & 0.038 & 0.080 & 0.037 & 0.057 & 0.048 & 0.026 & 0.031 & 0.071 & 0.009 & 0.014 & 0.073 & 0.022 & 0.029 \\ 
\sgns & \sgns & \lstm & \none & $450$ & 0.075 & 0.028 & 0.035 & 0.082 & 0.038 & 0.056 & 0.052 & 0.025 & 0.030 & 0.070 & 0.015 & 0.009 & 0.073 & 0.022 & 0.030 \\ 
\submitrow
\allvec & \sgns & \gru & \none & $450$ & 0.093 & 0.033 & 0.042 & 0.089 & 0.044 & 0.062 & 0.049 & \bsts{0.028} & \bsts{0.033} & 0.076 & \bsts{0.010} & \bsts{0.015} & 0.075 & \bsts{0.027} & \bsts{0.036} \\ 
\submitrow
\allvec & \sgns & \lstm & \none & $450$ & 0.091 & 0.033 & 0.041 & 0.092 & 0.044 & 0.061 & 0.051 & 0.027 & 0.032 & 0.076 & 0.010 & 0.015 & 0.076 & 0.026 & 0.033 \\ 
\allvec & \sgns & \gru & \sgns & $450$ & \bst{0.096} & \bst{0.033} & \bst{0.042} & \bst{0.096} & \bst{0.044} & 0.063 & \bst{0.061} & \bst{0.028} & \bst{0.033} & \bst{0.077} & \bst{0.010} & \bst{0.015} & \bst{0.082} & \bst{0.028} & \bst{0.037} \\ 
\allvec & \sgns & \lstm & \sgns & $450$ & 0.095 & 0.033 & 0.042 & 0.093 & 0.044 & 0.061 & 0.057 & 0.027 & 0.032 & 0.077 & 0.010 & 0.015 & 0.079 & 0.026 & 0.034 \\ 
\submitrow
\allvec & \sgns & \lstm & \gru & $450$ & \bsts{0.094} & \bsts{0.033} & \bsts{0.042} & 0.092 & 0.044 & 0.061 & \bsts{0.056} & 0.027 & 0.032 & \bsts{0.077} & 0.010 & 0.015 & 0.078 & 0.026 & 0.033 \\ 
\midrule
\electra & \electra & \gru & \none & $450$ & 0.079 & 0.028 & 0.034 & ~ & ~ & ~ & 0.050 & 0.029 & 0.024 & ~ & ~ & ~ & 0.072 & 0.024 & 0.031 \\ 
\allvec & \electra & \gru & \none & $450$ & 0.091 & 0.032 & 0.041 & ~ & ~ & ~ & 0.047 & 0.027 & 0.032 & ~ & ~ & ~ & 0.073 & 0.027 & 0.035 \\ 
\electra & \electra & \gru & \sgns & $450$ & 0.094 & 0.033 & 0.041 & ~ & ~ & ~ & 0.058 & 0.026 & 0.031 & ~ & ~ & ~ & 0.082 & 0.028 & 0.036 \\ 
\bottomrule
\end{tabular}
\label{table:defmoddetails}
\end{table} 
\end{landscape}
}

\subsection{Qualitative Analysis of Generated Glosses}
\label{app:gloss-analysis}

The \defmod models achieve weak results in comparison to the previous
state-of-art approaches, which is probably due to the comparably small amount of training and pretraining data. Here we demonstrate that the generated glosses can nevertheless capture a degree of the semantics of the correct glosses.

Table \ref{table:glosses1} shows four categories of semantic similarity between the correct and model-generated glosses, in descending order (highest similarity first).  These categories include hits or near hits (correct glosses), ``near misses'' (glosses that capture a significant amount of the original meaning), somewhat similar glosses, and complete misses. Several examples demonstrate that the subword-based models can produce syntactically incorrect glosses.

Table \ref{table:glosses2} contains generated glosses for different senses of the word ``consider'', which demonstrate that the model was able to approximate, to a degree, the semantics of the senses.

A principled analysis of the generated and correct glosses, based on a well defined semantic annotation scheme, might prove revealing but it would be time-consuming and impractical. Therefore it would be of interest to automatize such efforts. It would be interesting to explore if this can be done using
large pretrained transformers able to measure fine-grained semantic similarity.

\newcommand{\gngls}[1]{\texttt{#1}}

{
\setlength{\tabcolsep}{0.26em}
\begin{table*}[h!]
\caption{Glosses generated by the top submitted \defmod model, 
alongside the correct glosses. The examples are ordered 
by descending semantic similarity between the correct and the generated gloss.}
\centering
\normalsize
\begin{tabular}{cl}
\toprule
 Word & True Gloss / \gngls{Generated Gloss} \\
\midrule
lamebrain & A fool \\
 &  \gngls{A fool , idiot} \\
sentiment & A general thought , feeling , or sense \\
 &  \gngls{A feeling or feeling of thinking} \\
available & Such as one may avail oneself of ; capable of being used for the accomplishment of a purpose \\
 & \gngls{Able to be used} \\
\midrule
model & A representation of a physical object , usually in miniature \\
 & \gngls{An act of designing} \\
 supernumerary & Of an organ or structure : additional to what is normally present \\
 & \gngls{Having four wings} \\
 navy & Belonging to the navy ; typical of the navy \\
 & \gngls{To be armed} \\
\midrule
 fuzzy & Vague or imprecise \\
 & \gngls{lacking} \\
  co-opt & To absorb or assimilate into an established group \\
 & \gngls{To conceal} \\
  misinformation & Information that is incorrect \\
 & \gngls{prejudice} \\
\midrule
  cutthroat & Ruthlessly competitive , dog-eat-dog \\
 & \gngls{Very large} \\
  discretional & discretionary \\
 & \gngls{Of or pertaining to} \\
  abundantly & In an abundant manner ; in a sufficient degree ; in large measure \\
 & \gngls{In a very manner} \\
\bottomrule
\end{tabular}
\label{table:glosses1}
\end{table*} 
}

{
\setlength{\tabcolsep}{0.26em}
\begin{table*}[h!]
Glosses generated by the top submitted \defmod model, 
alongside the correct glosses, for the multiple senses of the word ``consider''.
\caption{}
\centering
\normalsize
\begin{tabular}{cl}
\toprule
 Word & True Gloss (describing the sense) / \gngls{Generated Gloss} \\
\midrule
consider & To assign some quality to \\
 &  \gngls{To hold the opinion} \\
consider & To look at attentively \\
 &  \gngls{To make something certain} \\
consider & To have regard to ; to take into view or account ; to pay due attention to ; to respect \\
 & \gngls{To hold into} \\
consider & To think of doing \\
 & \gngls{To permit} \\
 consider & To debate ( or dispose of ) a motion \\
  & \gngls{To make something certain} \\
\bottomrule
\end{tabular}
\label{table:glosses2}
\end{table*} 
}

\section{Appendix - Analysis of \revdict Data and Models}

\subsection{Data Analysis}
\label{appdx:revdict-data-analysis}
\begin{figure*}
    \centering
    \includegraphics[width=\linewidth]{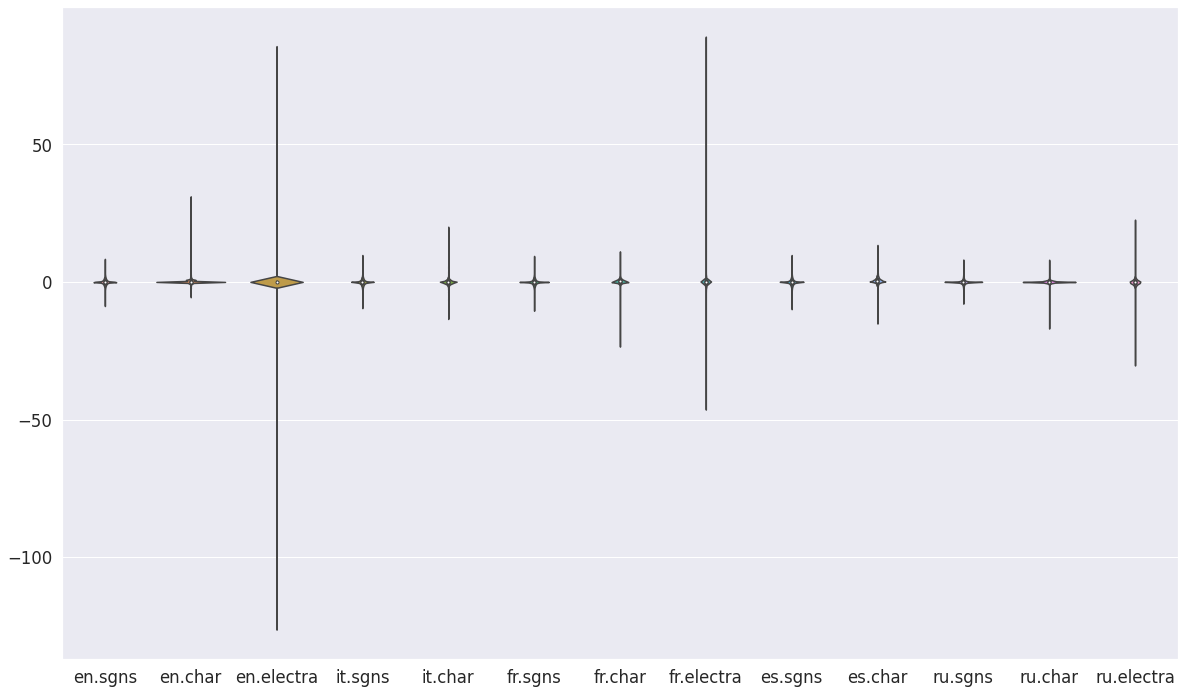}
    \caption{Distributions of vector elements in train datasets.}
    \label{fig:dist-train-vec-full}
\end{figure*}

\begin{figure*}
    \centering
    \includegraphics[width=\linewidth]{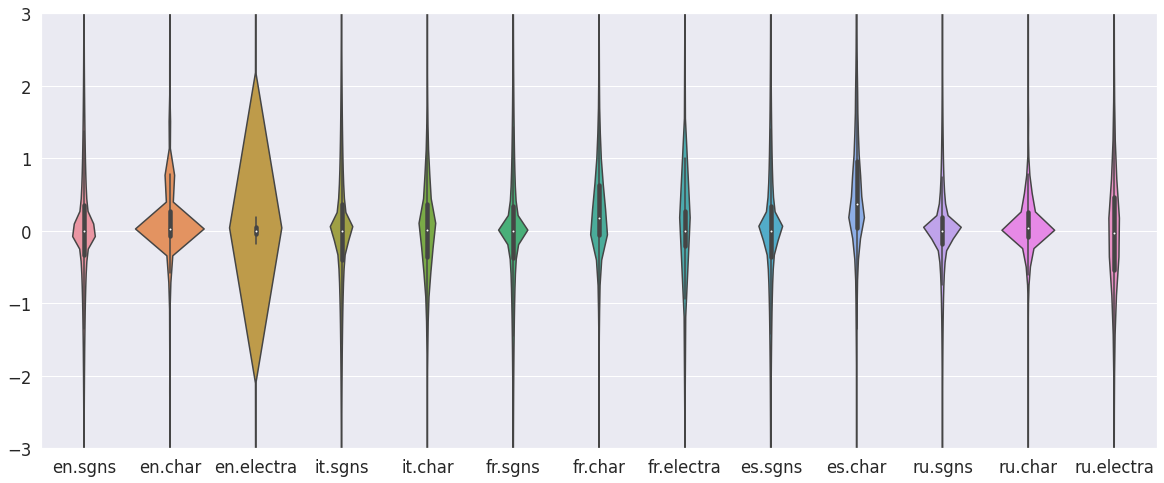}
    \caption{Distributions of vector elements in train datasets within the interval [-3,3].}
    \label{fig:dist-train-vec}
\end{figure*}

Here we analyze the properties of the pretrained embedding vectors assigned to the words defined by the glosses. We start by analyzing the numeric values contained in the vectors.
Basic statistics of vector elements can be found in Table \ref{tab:train-data-stats}. It is noticeable that there are large variations in value depending on the language and the embedding type. For example, there is a significant difference between maximum values, especially between \electra and \sgns.
To further investigate the vector elements, we visualize the shapes of their distributions for train datasets (Figure \ref{fig:dist-train-vec-full} and \ref{fig:dist-train-vec}). Distribution shapes look similar for dev datasets.

\begin{figure*}
    \centering
    \includegraphics[width=\linewidth]{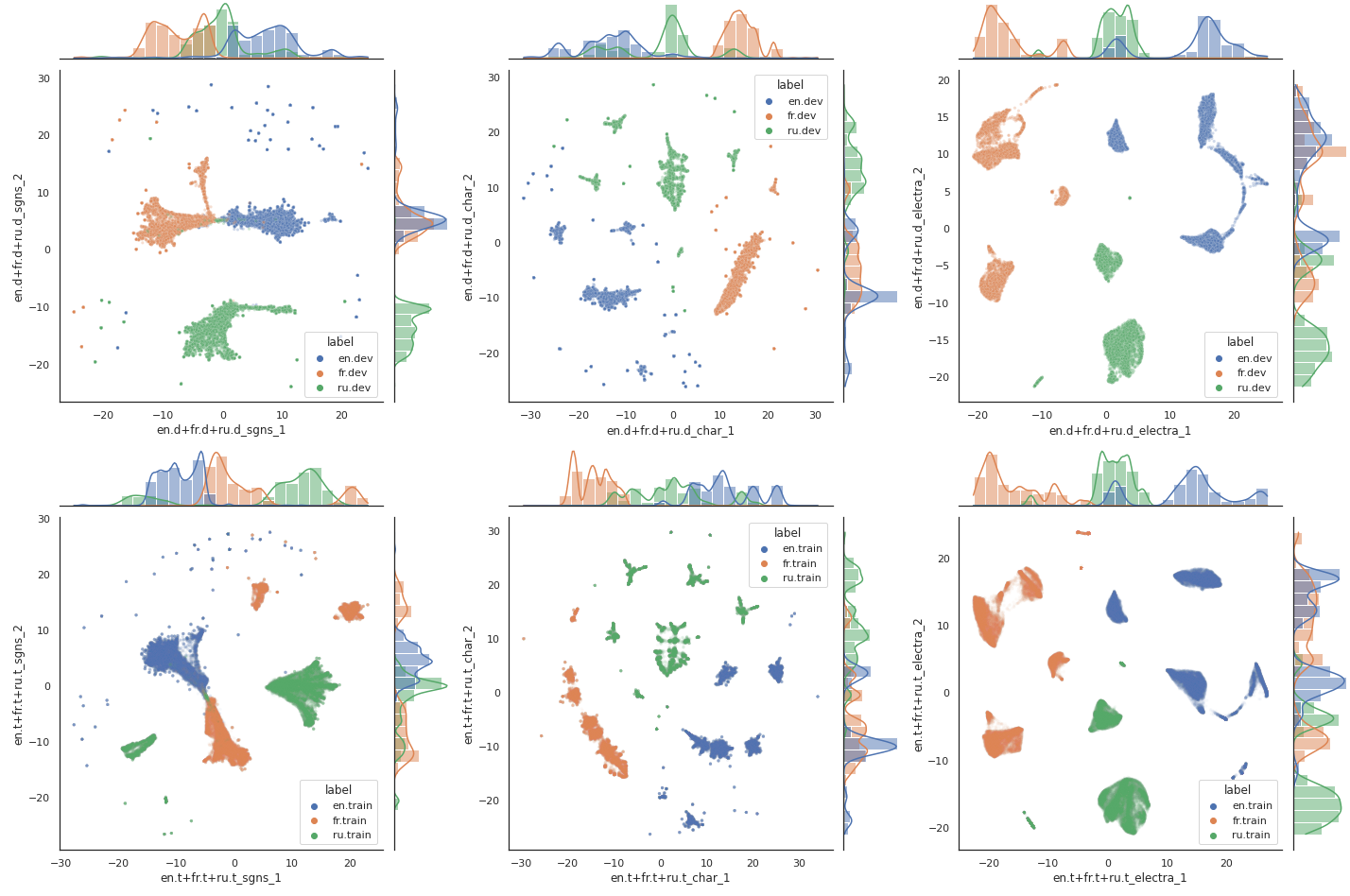}
    \caption{Distributions of all three embedding types in train (2nd row) and validation (development, 1st row) datasets after dimensionality reduction to 2D space. 
    \sgns (word2vec, 1st column), \chr (2nd column), and \electra (3rd column) embeddings are depicted for English (orange), French (green), and Russian (blue).}
    \label{fig:dist-vec-3e-3l-2D}
\end{figure*}

\begin{figure*}
    \centering
    \includegraphics[width=\linewidth]{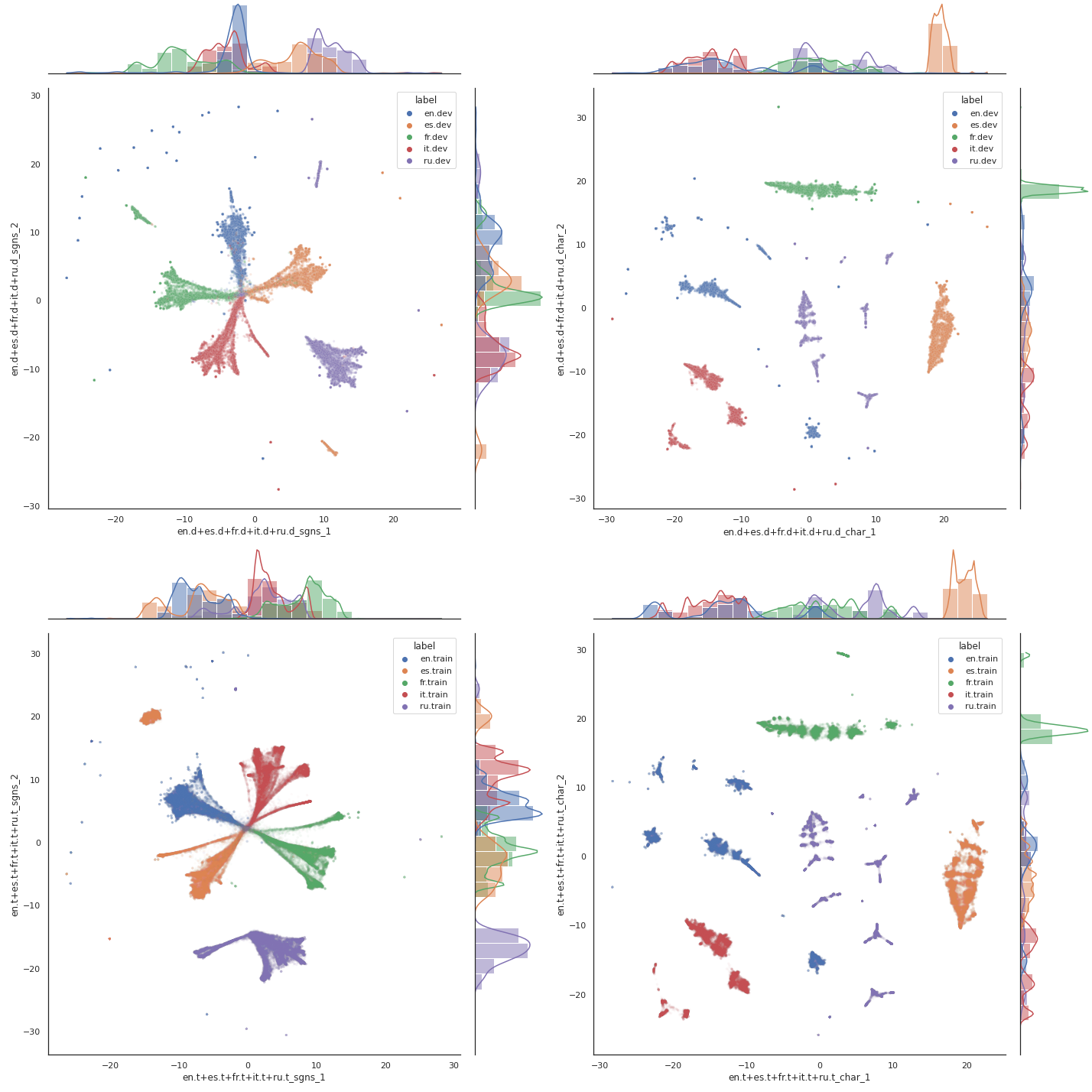}
    \caption{Distributions of \sgns (1st column) and \chr (2nd column) embeddings in train (2nd row) and validation (development, 1st row) datasets after dimensionality reduction to 2D space. The embeddings are depicted for English (blue), Spanish (orange), French (green), Italian (red), and Russian (violet).}
    \label{fig:dist-vec-2e-5l-2D}
\end{figure*}

Next, we explore the vector data by reducing dimensionality to the 2D space using the Pairwise Controlled Manifold Approximation Projection (PaCMAP) algorithm \cite{{wang-2020-pacmap}}. Figure \ref{fig:dist-vec-3e-3l-2D} shows the distributions of all three types of embeddings in the train and validation (development) datasets for English, French, and Russian. We also visualize distributions of \sgns (word2vec) and \chr embeddings for all languages, in Figure \ref{fig:dist-vec-2e-5l-2D}.
As can be seen, the vector distributions vary greatly between the embedding types. Additionally, for all the embedding types, the vectors of different languages occupy a distinct area and are easily separable.

\begin{figure*}
    \centering
    \includegraphics[width=\textwidth,height=\textheight,keepaspectratio]{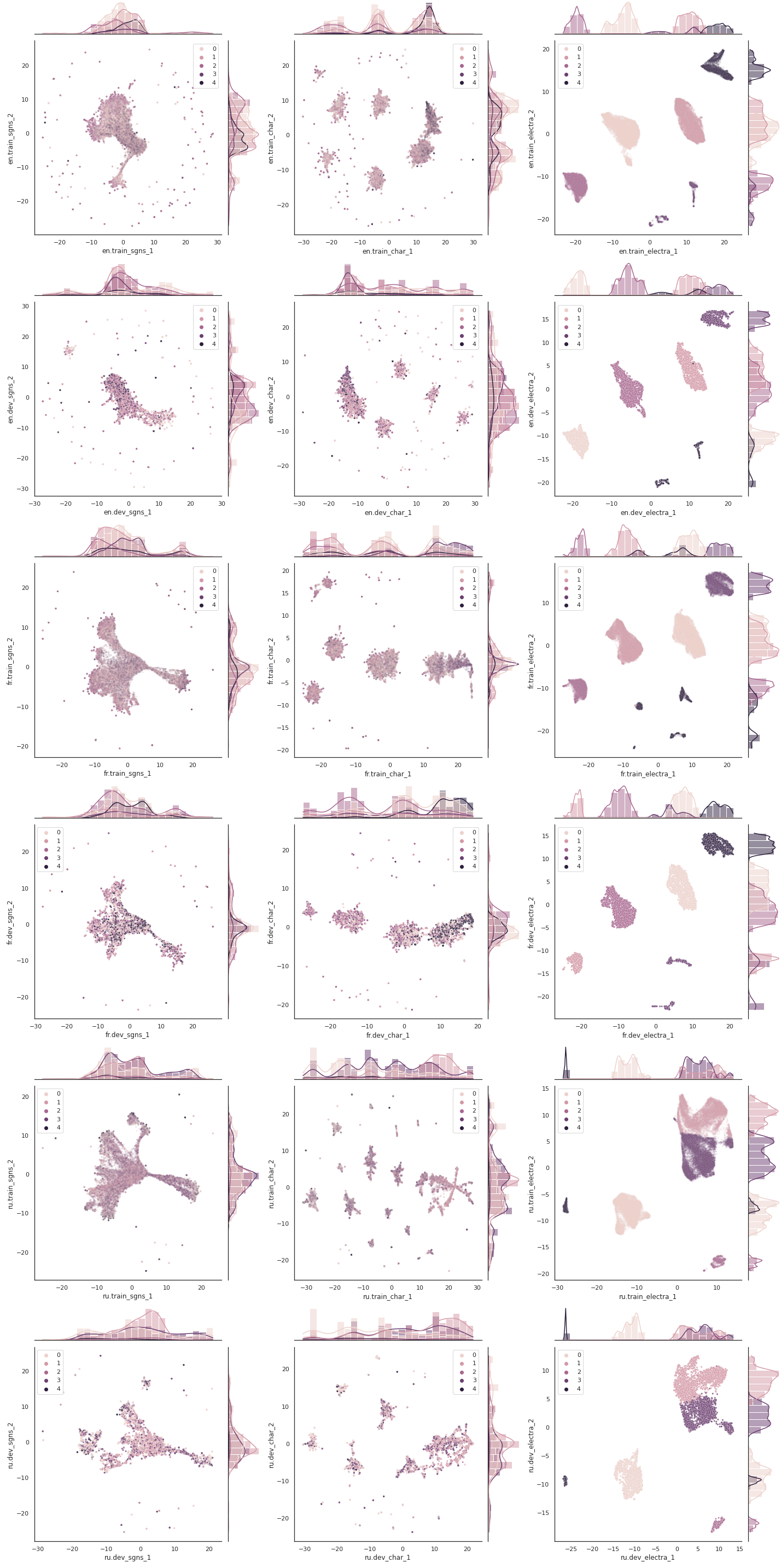}
    \caption{%
    Projection of the clusters in the \electra embedding space (3rd column) to the spaces of the other two embedding types \sgns (1st column) and \chr (2nd column). The analysis if performed for English (rows 1--2), French (rows 3--4), and Russian (rows 5--6) train and validation (development) datasets, after dimensionality reduction to 2D space.}
    \label{fig:dist-vec-3e-3l-2D-electra-clusters}
\end{figure*}

We further investigate the relationships between different embeddings in the following way. We first cluster the values of the \electra vectors with k-means algorithm. We set the number of clusters to five and assign a different color to each cluster. We retain the \electra cluster-based color of the samples (glosses) while visualizing the vectors of other embedding types, as shown in Figure \ref{fig:dist-vec-3e-3l-2D-electra-clusters}. It can be clearly seen that the \electra-based clusters are not preserved for other embedding types.

\subsection{Model Performance}
\label{appdx:revdict-model-analysis}
Here we present validation and test scores for our six \revdict solutions described in Section \ref{sect:revdict-experimental}. We use the following metrics for internal validation of our \revdict solutions: \emph{Mean Squared Error} (MSE), \emph{Cosine Similarity} (COS) and \emph{Central Kernel Alignment} (CKA) \cite{kornblith-2019-cka, cortes-2012-cka}. Validation scores for each \revdict approach can be found in Table \ref{tab:rd-val-score-all}.
The last three rows contain the total scores for each metric and each of our \revdict solutions. A total score is the sum of the values of all datasets and we use it for a simple comparison of solutions.
It is evident that each subsequent approach gives better validation results than the previous ones.

Test predictions are scored by these metrics: MSE, COS, and \emph{Cosine-Based Ranking} (RNK). The RNK measure is defined as the proportion of test samples with cosine similarity to the model output embedding higher than the ground truth embedding. The final results for all our solutions can be found in Table \ref{tab:rd-test-score-all}. Here, each subsequent approach has lower scores than the previous ones, which is the complete opposite of the validation results. This suggests potential overfitting to the dev dataset that could be the result of Bayesian hyperparameter optimization (BHO). However, this is contrary to expectations as the last two solutions have three times fewer BHO points and should not overfit to the dev dataset. The reason for this phenomenon is unclear and needs further investigation.

The best \revdict results for each team can be found in Table \ref{tab:rd-test-mse-all-teams} for MSE score, Table \ref{tab:rd-test-cos-all-teams} for COS score, and Table \ref{tab:rd-test-rnk-all-teams} for RNK score.
When compared to other solutions, our systems have low to average performance according to the MSE scores. For the COS scores, our systems have very good performance on \sgns (word2vec) vectors, and low performance on other embedding types. In terms of the RNK (ranking) our systems almost always yield the top performance, and this result is consistent across languages and embedding types.

\begin{table*}
\centering
\begin{tabular}{lllrrrrrr}
\hline
\textbf{lang} &  \textbf{split} &   \textbf{vector} &         \textbf{min} &      \textbf{mean} &       \textbf{max} &      \textbf{abs-min} &  \textbf{abs-mean} &    \textbf{abs-max} \\
\hline
  en &  train &     sgns &   -8.66 &  0.012 &  8.33 & 2.40-08 &  0.641 &   8.66 \\
  en &  train &     char &   -5.48 &  0.081 & 31.10 & 6.60-09 &  0.341 &  31.10 \\
  en &  train &  electra & -126.26 &  0.033 & 85.62 & 1.00-10 &  0.598 & 126.26 \\
  en &    dev &     sgns &   -7.02 &  0.013 &  7.30 & 9.51-08 &  0.657 &   7.30 \\
  en &    dev &     char &   -5.48 &  0.083 &  7.31 & 8.75-08 &  0.341 &   7.31 \\
  en &    dev &  electra &  -48.24 &  0.028 & 52.19 & 1.00-09 &  0.587 &  52.19 \\
\hline
  it &  train &     sgns &   -9.41 & -0.014 &  9.72 & 6.60-09 &  0.700 &   9.72 \\
  it &  train &     char &  -13.37 &  0.013 & 20.02 & 1.62-07 &  0.553 &  20.02 \\
  it &    dev &     sgns &   -8.22 & -0.013 &  7.82 & 1.02-07 &  0.706 &   8.22 \\
  it &    dev &     char &   -9.95 &  0.008 & 16.23 & 3.96-07 &  0.551 &  16.23 \\
\hline
  fr &  train &     sgns &  -10.38 & -0.013 &  9.39 & 1.59-08 &  0.682 &  10.38 \\
  fr &  train &     char &  -23.42 &  0.306 & 11.07 & 1.12-08 &  0.574 &  23.42 \\
  fr &  train &  electra &  -46.24 &  0.045 & 89.07 & 3.00-10 &  0.644 &  89.07 \\
  fr &    dev &     sgns &   -7.57 & -0.017 &  7.81 & 3.51-07 &  0.666 &   7.81 \\
  fr &    dev &     char &  -14.60 &  0.307 &  7.80 & 2.62-07 &  0.574 &  14.60 \\
  fr &    dev &  electra &  -42.73 &  0.045 & 51.29 & 9.00-10 &  0.655 &  51.29 \\
\hline
  es &  train &     sgns &   -9.79 & -0.018 &  9.72 & 2.15-08 &  0.653 &   9.79 \\
  es &  train &     char &  -15.03 &  0.577 & 13.37 & 2.27-07 &  0.822 &  15.03 \\
  es &    dev &     sgns &   -9.32 & -0.021 &  7.22 & 8.86-08 &  0.658 &   9.32 \\
  es &    dev &     char &  -13.19 &  0.577 & 11.40 & 2.28-06 &  0.820 &  13.19 \\
\hline
  ru &  train &     sgns &   -7.82 &  0.002 &  8.08 & 1.17-07 &  0.446 &   8.08 \\
  ru &  train &     char &  -16.87 &  0.139 &  8.04 & 8.00-10 &  0.311 &  16.87 \\
  ru &  train &  electra &  -30.24 & -0.017 & 22.56 & 1.75-08 &  0.788 &  30.24 \\
  ru &    dev &     sgns &   -8.06 &  0.002 &  7.91 & 7.94-08 &  0.439 &   8.06 \\
  ru &    dev &     char &  -11.86 &  0.140 &  8.01 & 3.05-07 &  0.310 &  11.86 \\
  ru &    dev &  electra &  -22.53 & -0.017 & 21.70 & 4.75-08 &  0.789 &  22.53 \\
\hline
\end{tabular}
\caption{\label{tab:train-data-stats} %
Statistics of the elements of the embedding vectors from the train and validation (development) datasets.}
\end{table*}

\begin{table*}[]
\centering
\begin{tabular}{l|llllll|l}
\hline
\textbf{METRICS} & \textbf{RD 1} & \textbf{RD 2} & \textbf{RD 3} & \textbf{RD 4} & \textbf{RD 5} & \textbf{RD 6} & \textbf{BEST} \\
\hline
\textbf{mse-en-sgns}    & \cellcolor[HTML]{FCB079}0.521           & \cellcolor[HTML]{F8696B}0.632           & \cellcolor[HTML]{FCB079}0.521           & \cellcolor[HTML]{FFEB84}0.428           & \cellcolor[HTML]{6CC07B}0.348           & \cellcolor[HTML]{63BE7B}0.343           & \cellcolor[HTML]{63BE7B}\textbf{0.343}  \\
\textbf{mse-en-char}    & \cellcolor[HTML]{FFEB84}0.088           & \cellcolor[HTML]{FECA7E}0.091           & \cellcolor[HTML]{63BE7B}0.051           & \cellcolor[HTML]{F8696B}0.098           & \cellcolor[HTML]{80C67C}0.058           & \cellcolor[HTML]{FED280}0.090           & \cellcolor[HTML]{63BE7B}\textbf{0.051}  \\
\textbf{mse-en-electra} & \cellcolor[HTML]{FDB67A}0.611           & \cellcolor[HTML]{F8696B}0.683           & \cellcolor[HTML]{FDB47A}0.612           & \cellcolor[HTML]{FFEB84}0.560           & \cellcolor[HTML]{B7D67F}0.439           & \cellcolor[HTML]{63BE7B}0.295           & \cellcolor[HTML]{63BE7B}\textbf{0.295}  \\
\textbf{mse-it-sgns}    & \cellcolor[HTML]{F8696B}0.846           & \cellcolor[HTML]{FED580}0.700           & \cellcolor[HTML]{FB9874}0.783           & \cellcolor[HTML]{EEE683}0.650           & \cellcolor[HTML]{FFEB84}0.670           & \cellcolor[HTML]{63BE7B}0.485           & \cellcolor[HTML]{63BE7B}\textbf{0.485}  \\
\textbf{mse-it-char}    & \cellcolor[HTML]{F8696B}0.264           & \cellcolor[HTML]{FFEB84}0.235           & \cellcolor[HTML]{FB9C75}0.253           & \cellcolor[HTML]{FA8571}0.258           & \cellcolor[HTML]{7AC47C}0.224           & \cellcolor[HTML]{63BE7B}0.222           & \cellcolor[HTML]{63BE7B}\textbf{0.222}  \\
\textbf{mse-fr-sgns}    & \cellcolor[HTML]{F8696B}0.773           & \cellcolor[HTML]{FCB079}0.671           & \cellcolor[HTML]{FECD7F}0.629           & \cellcolor[HTML]{FFEB84}0.585           & \cellcolor[HTML]{B1D47F}0.514           & \cellcolor[HTML]{63BE7B}0.443           & \cellcolor[HTML]{63BE7B}\textbf{0.443}  \\
\textbf{mse-fr-char}    & \cellcolor[HTML]{FFDB81}0.210           & \cellcolor[HTML]{FFEB84}0.205           & \cellcolor[HTML]{FED781}0.211           & \cellcolor[HTML]{F8696B}0.246           & \cellcolor[HTML]{6EC17B}0.180           & \cellcolor[HTML]{63BE7B}0.178           & \cellcolor[HTML]{63BE7B}\textbf{0.178}  \\
\textbf{mse-fr-electra} & \cellcolor[HTML]{F97A6F}0.634           & \cellcolor[HTML]{FA8C72}0.616           & \cellcolor[HTML]{FFEB84}0.518           & \cellcolor[HTML]{F8696B}0.651           & \cellcolor[HTML]{8AC97D}0.400           & \cellcolor[HTML]{63BE7B}0.360           & \cellcolor[HTML]{63BE7B}\textbf{0.360}  \\
\textbf{mse-es-sgns}    & \cellcolor[HTML]{FDC27C}0.627           & \cellcolor[HTML]{F8696B}0.764           & \cellcolor[HTML]{FCA376}0.675           & \cellcolor[HTML]{EEE683}0.560           & \cellcolor[HTML]{63BE7B}0.543           & \cellcolor[HTML]{FFEB84}0.562           & \cellcolor[HTML]{63BE7B}\textbf{0.543}  \\
\textbf{mse-es-char}    & \cellcolor[HTML]{FB9D75}0.341           & \cellcolor[HTML]{FCAA78}0.338           & \cellcolor[HTML]{FFEB84}0.323           & \cellcolor[HTML]{F8696B}0.353           & \cellcolor[HTML]{A8D27F}0.291           & \cellcolor[HTML]{63BE7B}0.265           & \cellcolor[HTML]{63BE7B}\textbf{0.265}  \\
\textbf{mse-ru-sgns}    & \cellcolor[HTML]{F8696B}0.363           & \cellcolor[HTML]{FED380}0.236           & \cellcolor[HTML]{FDB97B}0.268           & \cellcolor[HTML]{FFEB84}0.207           & \cellcolor[HTML]{B7D67F}0.162           & \cellcolor[HTML]{63BE7B}0.109           & \cellcolor[HTML]{63BE7B}\textbf{0.109}  \\
\textbf{mse-ru-char}    & \cellcolor[HTML]{FFEB84}0.053           & \cellcolor[HTML]{FA8471}0.060           & \cellcolor[HTML]{F8696B}0.062           & \cellcolor[HTML]{FA8671}0.060           & \cellcolor[HTML]{63BE7B}0.038           & \cellcolor[HTML]{EAE482}0.051           & \cellcolor[HTML]{63BE7B}\textbf{0.038}  \\
\textbf{mse-ru-electra} & \cellcolor[HTML]{F8696B}0.544           & \cellcolor[HTML]{FFEB84}0.481           & \cellcolor[HTML]{FB9D75}0.519           & \cellcolor[HTML]{FECB7E}0.497           & \cellcolor[HTML]{63BE7B}0.313           & \cellcolor[HTML]{C7DA80}0.421           & \cellcolor[HTML]{63BE7B}\textbf{0.313}  \\
\hline
\textbf{cos-en-sgns}    & \cellcolor[HTML]{FA9773}0.483           & \cellcolor[HTML]{F8696B}0.453           & \cellcolor[HTML]{FBA576}0.492           & \cellcolor[HTML]{FFEB84}0.537           & \cellcolor[HTML]{63BE7B}0.571           & \cellcolor[HTML]{C8DC81}0.549           & \cellcolor[HTML]{63BE7B}\textbf{0.571}  \\
\textbf{cos-en-char}    & \cellcolor[HTML]{FFEB84}0.875           & \cellcolor[HTML]{FDC87D}0.871           & \cellcolor[HTML]{63BE7B}0.927           & \cellcolor[HTML]{F8696B}0.860           & \cellcolor[HTML]{7EC67D}0.918           & \cellcolor[HTML]{FED980}0.873           & \cellcolor[HTML]{63BE7B}\textbf{0.927}  \\
\textbf{cos-en-electra} & \cellcolor[HTML]{FCB97A}0.895           & \cellcolor[HTML]{F8696B}0.887           & \cellcolor[HTML]{FCC37C}0.896           & \cellcolor[HTML]{FFEB84}0.900           & \cellcolor[HTML]{C2DA81}0.915           & \cellcolor[HTML]{63BE7B}0.938           & \cellcolor[HTML]{63BE7B}\textbf{0.938}  \\
\textbf{cos-it-sgns}    & \cellcolor[HTML]{F8696B}0.470           & \cellcolor[HTML]{FDCE7E}0.510           & \cellcolor[HTML]{F98770}0.482           & \cellcolor[HTML]{F0E784}0.527           & \cellcolor[HTML]{FFEB84}0.521           & \cellcolor[HTML]{63BE7B}0.580           & \cellcolor[HTML]{63BE7B}\textbf{0.580}  \\
\textbf{cos-it-char}    & \cellcolor[HTML]{F8696B}0.801           & \cellcolor[HTML]{FFEB84}0.824           & \cellcolor[HTML]{FA9B74}0.810           & \cellcolor[HTML]{F98A71}0.807           & \cellcolor[HTML]{7DC67D}0.834           & \cellcolor[HTML]{63BE7B}0.836           & \cellcolor[HTML]{63BE7B}\textbf{0.836}  \\
\textbf{cos-fr-sgns}    & \cellcolor[HTML]{F8696B}0.457           & \cellcolor[HTML]{FBAA77}0.490           & \cellcolor[HTML]{FDCE7E}0.508           & \cellcolor[HTML]{FFEB84}0.522           & \cellcolor[HTML]{9BCE7F}0.544           & \cellcolor[HTML]{63BE7B}0.556           & \cellcolor[HTML]{63BE7B}\textbf{0.556}  \\
\textbf{cos-fr-char}    & \cellcolor[HTML]{FDD780}0.866           & \cellcolor[HTML]{FFEB84}0.870           & \cellcolor[HTML]{FDD780}0.866           & \cellcolor[HTML]{F8696B}0.843           & \cellcolor[HTML]{6DC17C}0.886           & \cellcolor[HTML]{63BE7B}0.887           & \cellcolor[HTML]{63BE7B}\textbf{0.887}  \\
\textbf{cos-fr-electra} & \cellcolor[HTML]{F97E6F}0.894           & \cellcolor[HTML]{F97E6F}0.894           & \cellcolor[HTML]{FFEB84}0.904           & \cellcolor[HTML]{F8696B}0.892           & \cellcolor[HTML]{95CD7E}0.921           & \cellcolor[HTML]{63BE7B}0.929           & \cellcolor[HTML]{63BE7B}\textbf{0.929}  \\
\textbf{cos-es-sgns}    & \cellcolor[HTML]{FCBA7A}0.501           & \cellcolor[HTML]{F8696B}0.450           & \cellcolor[HTML]{FBA276}0.486           & \cellcolor[HTML]{FFEB84}0.531           & \cellcolor[HTML]{63BE7B}0.539           & \cellcolor[HTML]{77C47D}0.538           & \cellcolor[HTML]{63BE7B}\textbf{0.539}  \\
\textbf{cos-es-char}    & \cellcolor[HTML]{FA9473}0.881           & \cellcolor[HTML]{FBB178}0.883           & \cellcolor[HTML]{FFEB84}0.887           & \cellcolor[HTML]{F8696B}0.878           & \cellcolor[HTML]{A6D27F}0.899           & \cellcolor[HTML]{63BE7B}0.908           & \cellcolor[HTML]{63BE7B}\textbf{0.908}  \\
\textbf{cos-ru-sgns}    & \cellcolor[HTML]{F8696B}0.525           & \cellcolor[HTML]{FDD07E}0.599           & \cellcolor[HTML]{FCB87A}0.582           & \cellcolor[HTML]{FFEB84}0.618           & \cellcolor[HTML]{85C87D}0.662           & \cellcolor[HTML]{63BE7B}0.674           & \cellcolor[HTML]{63BE7B}\textbf{0.674}  \\
\textbf{cos-ru-char}    & \cellcolor[HTML]{FFEB84}0.933           & \cellcolor[HTML]{F98C71}0.925           & \cellcolor[HTML]{F8696B}0.922           & \cellcolor[HTML]{F98C71}0.925           & \cellcolor[HTML]{63BE7B}0.952           & \cellcolor[HTML]{EFE784}0.935           & \cellcolor[HTML]{63BE7B}\textbf{0.952}  \\
\textbf{cos-ru-electra} & \cellcolor[HTML]{F8696B}0.807           & \cellcolor[HTML]{FFEB84}0.821           & \cellcolor[HTML]{FA8E72}0.811           & \cellcolor[HTML]{FCC57C}0.817           & \cellcolor[HTML]{63BE7B}0.872           & \cellcolor[HTML]{C2DA81}0.841           & \cellcolor[HTML]{63BE7B}\textbf{0.872}  \\
\hline
\textbf{cka-en-sgns}    & \cellcolor[HTML]{F8696B}0.755           & \cellcolor[HTML]{F97D6E}0.776           & \cellcolor[HTML]{FEE182}0.879           & \cellcolor[HTML]{63BE7B}0.939           & \cellcolor[HTML]{FFEB84}0.889           & \cellcolor[HTML]{EDE683}0.895           & \cellcolor[HTML]{63BE7B}\textbf{0.939}  \\
\textbf{cka-en-char}    & \cellcolor[HTML]{F8696B}0.991           & \cellcolor[HTML]{FCB77A}0.994           & \cellcolor[HTML]{63BE7B}0.998           & \cellcolor[HTML]{FFEB84}0.996           & \cellcolor[HTML]{FFEB84}0.996           & \cellcolor[HTML]{FCB77A}0.994           & \cellcolor[HTML]{63BE7B}\textbf{0.998}  \\
\textbf{cka-en-electra} & \cellcolor[HTML]{F8696B}0.993           & \cellcolor[HTML]{FBAA77}0.995           & \cellcolor[HTML]{FFEB84}0.997           & \cellcolor[HTML]{63BE7B}0.998           & \cellcolor[HTML]{FFEB84}0.997           & \cellcolor[HTML]{63BE7B}0.998           & \cellcolor[HTML]{63BE7B}\textbf{0.998}  \\
\textbf{cka-it-sgns}    & \cellcolor[HTML]{F8696B}0.608           & \cellcolor[HTML]{FDD27F}0.773           & \cellcolor[HTML]{FFEB84}0.811           & \cellcolor[HTML]{63BE7B}0.905           & \cellcolor[HTML]{FDD780}0.780           & \cellcolor[HTML]{B3D580}0.857           & \cellcolor[HTML]{63BE7B}\textbf{0.905}  \\
\textbf{cka-it-char}    & \cellcolor[HTML]{F8696B}0.979           & \cellcolor[HTML]{FEDF81}0.989           & \cellcolor[HTML]{98CE7F}0.992           & \cellcolor[HTML]{63BE7B}0.993           & \cellcolor[HTML]{FEDF81}0.989           & \cellcolor[HTML]{FFEB84}0.990           & \cellcolor[HTML]{63BE7B}\textbf{0.993}  \\
\textbf{cka-fr-sgns}    & \cellcolor[HTML]{F8696B}0.647           & \cellcolor[HTML]{FCB97A}0.779           & \cellcolor[HTML]{FFEB84}0.860           & \cellcolor[HTML]{63BE7B}0.916           & \cellcolor[HTML]{FEDD81}0.838           & \cellcolor[HTML]{EFE784}0.866           & \cellcolor[HTML]{63BE7B}\textbf{0.916}  \\
\textbf{cka-fr-char}    & \cellcolor[HTML]{F8696B}0.981           & \cellcolor[HTML]{FDD17F}0.989           & \cellcolor[HTML]{63BE7B}0.993           & \cellcolor[HTML]{FFEB84}0.991           & \cellcolor[HTML]{FEDE81}0.990           & \cellcolor[HTML]{FFEB84}0.991           & \cellcolor[HTML]{63BE7B}\textbf{0.993}  \\
\textbf{cka-fr-electra} & \cellcolor[HTML]{F8696B}0.991           & \cellcolor[HTML]{FCBF7B}0.995           & \cellcolor[HTML]{63BE7B}0.997           & \cellcolor[HTML]{FDD57F}0.996           & \cellcolor[HTML]{63BE7B}0.997           & \cellcolor[HTML]{63BE7B}0.997           & \cellcolor[HTML]{63BE7B}\textbf{0.997}  \\
\textbf{cka-es-sgns}    & \cellcolor[HTML]{F8696B}0.685           & \cellcolor[HTML]{F8776D}0.698           & \cellcolor[HTML]{DDE182}0.826           & \cellcolor[HTML]{63BE7B}0.909           & \cellcolor[HTML]{FFEB84}0.802           & \cellcolor[HTML]{FEE382}0.795           & \cellcolor[HTML]{63BE7B}\textbf{0.909}  \\
\textbf{cka-es-char}    & \cellcolor[HTML]{F8696B}0.982           & \cellcolor[HTML]{FCC47C}0.989           & \cellcolor[HTML]{63BE7B}0.993           & \cellcolor[HTML]{FFEB84}0.992           & \cellcolor[HTML]{FEDE81}0.991           & \cellcolor[HTML]{FFEB84}0.992           & \cellcolor[HTML]{63BE7B}\textbf{0.993}  \\
\textbf{cka-ru-sgns}    & \cellcolor[HTML]{F8696B}0.611           & \cellcolor[HTML]{FDCE7E}0.821           & \cellcolor[HTML]{FEDC81}0.850           & \cellcolor[HTML]{63BE7B}0.927           & \cellcolor[HTML]{FFEB84}0.880           & \cellcolor[HTML]{63BE7B}0.927           & \cellcolor[HTML]{63BE7B}\textbf{0.927}  \\
\textbf{cka-ru-char}    & \cellcolor[HTML]{F8696B}0.995           & \cellcolor[HTML]{FFEB84}0.997           & \cellcolor[HTML]{63BE7B}0.998           & \cellcolor[HTML]{63BE7B}0.998           & \cellcolor[HTML]{FFEB84}0.997           & \cellcolor[HTML]{FFEB84}0.997           & \cellcolor[HTML]{63BE7B}\textbf{0.998}  \\
\textbf{cka-ru-electra} & \cellcolor[HTML]{F8696B}0.978           & \cellcolor[HTML]{FDCF7E}0.989           & \cellcolor[HTML]{FFEB84}0.992           & \cellcolor[HTML]{63BE7B}0.994           & \cellcolor[HTML]{63BE7B}0.994           & \cellcolor[HTML]{FEE182}0.991           & \cellcolor[HTML]{63BE7B}\textbf{0.994}  \\
\hline
\textbf{TOTAL   mse} & \cellcolor[HTML]{F8696B}\textbf{5.875}  & \cellcolor[HTML]{FA8771}\textbf{5.712}  & \cellcolor[HTML]{FDBB7B}\textbf{5.425}  & \cellcolor[HTML]{FFEB84}\textbf{5.153}  & \cellcolor[HTML]{8CCA7D}\textbf{4.180}  & \cellcolor[HTML]{63BE7B}\textbf{3.824}  & \cellcolor[HTML]{63BE7B}\textbf{3.824}  \\
\textbf{TOTAL cos}   & \cellcolor[HTML]{F8696B}\textbf{9.388}  & \cellcolor[HTML]{FA9473}\textbf{9.477}  & \cellcolor[HTML]{FCC27C}\textbf{9.573}  & \cellcolor[HTML]{FFEB84}\textbf{9.657}  & \cellcolor[HTML]{68C07C}\textbf{10.034} & \cellcolor[HTML]{63BE7B}\textbf{10.044} & \cellcolor[HTML]{63BE7B}\textbf{10.044} \\
\textbf{TOTAL cka}   & \cellcolor[HTML]{F8696B}\textbf{11.196} & \cellcolor[HTML]{FCB679}\textbf{11.784} & \cellcolor[HTML]{FFEB84}\textbf{12.186} & \cellcolor[HTML]{63BE7B}\textbf{12.554} & \cellcolor[HTML]{FEE482}\textbf{12.140} & \cellcolor[HTML]{D3DF82}\textbf{12.290} & \cellcolor[HTML]{63BE7B}\textbf{12.554} \\
\hline
\end{tabular}
\caption{\label{tab:rd-val-score-all} Validation scores for all our \revdict (RD) approaches. For each score, comparative results are shown in color. Green is used for the best and red for the worst-performing solution per row (a metric defines whether higher or lower values are better). The total score is the sum of the values over all datasets and embeddings.
}
\end{table*}

\begin{table*}
\centering
\begin{tabular}{llllllllllllll}
\hline
\textbf{METRICS} & \textbf{RD 1} & \textbf{RD 2} & \textbf{RD 3} & \textbf{RD 4} & \textbf{RD 5} & \textbf{RD 6} & \textbf{BEST} \\
\hline
\textbf{mse-en-sgns}    & \cellcolor[HTML]{FFEB84}1.024          & \cellcolor[HTML]{63BE7B}0.964          & \cellcolor[HTML]{F8E983}1.021          & \cellcolor[HTML]{FDB57A}1.085          & \cellcolor[HTML]{F8696B}1.170          & \cellcolor[HTML]{FB9674}1.119          & \cellcolor[HTML]{63BE7B}\textbf{0.964} \\
\textbf{mse-en-char}    & \cellcolor[HTML]{FFEB84}0.169          & \cellcolor[HTML]{FAE983}0.169          & \cellcolor[HTML]{FB9574}0.186          & \cellcolor[HTML]{63BE7B}0.162          & \cellcolor[HTML]{F8696B}0.195          & \cellcolor[HTML]{FFDD82}0.172          & \cellcolor[HTML]{63BE7B}\textbf{0.162} \\
\textbf{mse-en-electra} & \cellcolor[HTML]{FFEB84}1.723          & \cellcolor[HTML]{63BE7B}1.685          & \cellcolor[HTML]{7AC47C}1.690          & \cellcolor[HTML]{FED580}1.768          & \cellcolor[HTML]{FCA777}1.863          & \cellcolor[HTML]{F8696B}1.988          & \cellcolor[HTML]{63BE7B}\textbf{1.685} \\
\textbf{mse-it-sgns}    & \cellcolor[HTML]{63BE7B}1.076          & \cellcolor[HTML]{FFE884}1.160          & \cellcolor[HTML]{91CB7D}1.100          & \cellcolor[HTML]{FFEB84}1.156          & \cellcolor[HTML]{FDBF7C}1.211          & \cellcolor[HTML]{F8696B}1.318          & \cellcolor[HTML]{63BE7B}\textbf{1.076} \\
\textbf{mse-it-char}    & \cellcolor[HTML]{63BE7B}0.366          & \cellcolor[HTML]{FDC37D}0.383          & \cellcolor[HTML]{FFEB84}0.376          & \cellcolor[HTML]{94CC7D}0.370          & \cellcolor[HTML]{F8696B}0.399          & \cellcolor[HTML]{F96F6C}0.399          & \cellcolor[HTML]{63BE7B}\textbf{0.366} \\
\textbf{mse-fr-sgns}    & \cellcolor[HTML]{63BE7B}1.068          & \cellcolor[HTML]{DCE081}1.119          & \cellcolor[HTML]{FFEB84}1.134          & \cellcolor[HTML]{FFE283}1.147          & \cellcolor[HTML]{FB9A75}1.250          & \cellcolor[HTML]{F8696B}1.319          & \cellcolor[HTML]{63BE7B}\textbf{1.068} \\
\textbf{mse-fr-char}    & \cellcolor[HTML]{CFDD81}0.409          & \cellcolor[HTML]{FFE884}0.419          & \cellcolor[HTML]{FFEB84}0.418          & \cellcolor[HTML]{63BE7B}0.390          & \cellcolor[HTML]{F8696B}0.447          & \cellcolor[HTML]{FCA577}0.434          & \cellcolor[HTML]{63BE7B}\textbf{0.390} \\
\textbf{mse-fr-electra} & \cellcolor[HTML]{63BE7B}1.339          & \cellcolor[HTML]{A5D17E}1.347          & \cellcolor[HTML]{FEC97E}1.414          & \cellcolor[HTML]{FFEB84}1.358          & \cellcolor[HTML]{F9716D}1.554          & \cellcolor[HTML]{F8696B}1.566          & \cellcolor[HTML]{63BE7B}\textbf{1.339} \\
\textbf{mse-es-sgns}    & \cellcolor[HTML]{FFEB84}0.941          & \cellcolor[HTML]{63BE7B}0.883          & \cellcolor[HTML]{D0DD81}0.924          & \cellcolor[HTML]{FEC97E}0.965          & \cellcolor[HTML]{F8696B}1.031          & \cellcolor[HTML]{F97A6F}1.020          & \cellcolor[HTML]{63BE7B}\textbf{0.883} \\
\textbf{mse-es-char}    & \cellcolor[HTML]{63BE7B}0.526          & \cellcolor[HTML]{FFEB84}0.545          & \cellcolor[HTML]{FFEA84}0.546          & \cellcolor[HTML]{93CB7D}0.532          & \cellcolor[HTML]{FDB67A}0.582          & \cellcolor[HTML]{F8696B}0.635          & \cellcolor[HTML]{63BE7B}\textbf{0.526} \\
\textbf{mse-ru-sgns}    & \cellcolor[HTML]{63BE7B}0.568          & \cellcolor[HTML]{FFE683}0.604          & \cellcolor[HTML]{E7E482}0.596          & \cellcolor[HTML]{FFEB84}0.601          & \cellcolor[HTML]{F8696B}0.667          & \cellcolor[HTML]{FA8571}0.653          & \cellcolor[HTML]{63BE7B}\textbf{0.568} \\
\textbf{mse-ru-char}    & \cellcolor[HTML]{FFDE82}0.145          & \cellcolor[HTML]{83C77C}0.141          & \cellcolor[HTML]{FFEB84}0.142          & \cellcolor[HTML]{63BE7B}0.140          & \cellcolor[HTML]{F8696B}0.170          & \cellcolor[HTML]{FFE283}0.144          & \cellcolor[HTML]{63BE7B}\textbf{0.140} \\
\textbf{mse-ru-electra} & \cellcolor[HTML]{63BE7B}0.911          & \cellcolor[HTML]{D5DF81}0.944          & \cellcolor[HTML]{FFEB84}0.956          & \cellcolor[HTML]{FFE784}0.961          & \cellcolor[HTML]{F8696B}1.105          & \cellcolor[HTML]{FB9B75}1.049          & \cellcolor[HTML]{63BE7B}\textbf{0.911} \\
\hline
\textbf{cos-en-sgns}    & \cellcolor[HTML]{F6E984}0.250          & \cellcolor[HTML]{63BE7B}0.260          & \cellcolor[HTML]{FFEB84}0.250          & \cellcolor[HTML]{FEDB81}0.245          & \cellcolor[HTML]{FBA877}0.231          & \cellcolor[HTML]{F8696B}0.214          & \cellcolor[HTML]{63BE7B}\textbf{0.260} \\
\textbf{cos-en-char}    & \cellcolor[HTML]{FFEB84}0.761          & \cellcolor[HTML]{FEE983}0.761          & \cellcolor[HTML]{FA9273}0.743          & \cellcolor[HTML]{63BE7B}0.770          & \cellcolor[HTML]{F8696B}0.734          & \cellcolor[HTML]{C6DB81}0.765          & \cellcolor[HTML]{63BE7B}\textbf{0.770} \\
\textbf{cos-en-electra} & \cellcolor[HTML]{FFEB84}0.821          & \cellcolor[HTML]{63BE7B}0.828          & \cellcolor[HTML]{C5DB81}0.824          & \cellcolor[HTML]{FEDB80}0.818          & \cellcolor[HTML]{FCC07B}0.812          & \cellcolor[HTML]{F8696B}0.792          & \cellcolor[HTML]{63BE7B}\textbf{0.828} \\
\textbf{cos-it-sgns}    & \cellcolor[HTML]{63BE7B}0.380          & \cellcolor[HTML]{FEDB80}0.358          & \cellcolor[HTML]{B4D680}0.370          & \cellcolor[HTML]{FEE883}0.361          & \cellcolor[HTML]{FFEB84}0.361          & \cellcolor[HTML]{F8696B}0.339          & \cellcolor[HTML]{63BE7B}\textbf{0.380} \\
\textbf{cos-it-char}    & \cellcolor[HTML]{63BE7B}0.724          & \cellcolor[HTML]{FBA977}0.713          & \cellcolor[HTML]{FFEB84}0.717          & \cellcolor[HTML]{ABD380}0.721          & \cellcolor[HTML]{F8696B}0.709          & \cellcolor[HTML]{F98770}0.711          & \cellcolor[HTML]{63BE7B}\textbf{0.724} \\
\textbf{cos-fr-sgns}    & \cellcolor[HTML]{63BE7B}0.342          & \cellcolor[HTML]{CBDC81}0.336          & \cellcolor[HTML]{FFEB84}0.333          & \cellcolor[HTML]{FEE582}0.330          & \cellcolor[HTML]{FDD37F}0.319          & \cellcolor[HTML]{F8696B}0.255          & \cellcolor[HTML]{63BE7B}\textbf{0.342} \\
\textbf{cos-fr-char}    & \cellcolor[HTML]{D5DF82}0.744          & \cellcolor[HTML]{FEE482}0.738          & \cellcolor[HTML]{FFEB84}0.739          & \cellcolor[HTML]{63BE7B}0.756          & \cellcolor[HTML]{F8696B}0.725          & \cellcolor[HTML]{FCBF7B}0.734          & \cellcolor[HTML]{63BE7B}\textbf{0.756} \\
\textbf{cos-fr-electra} & \cellcolor[HTML]{63BE7B}0.847          & \cellcolor[HTML]{FFEB84}0.842          & \cellcolor[HTML]{FDC77D}0.837          & \cellcolor[HTML]{ADD480}0.844          & \cellcolor[HTML]{F9806F}0.828          & \cellcolor[HTML]{F8696B}0.825          & \cellcolor[HTML]{63BE7B}\textbf{0.847} \\
\textbf{cos-es-sgns}    & \cellcolor[HTML]{E4E483}0.362          & \cellcolor[HTML]{63BE7B}0.367          & \cellcolor[HTML]{FFEB84}0.361          & \cellcolor[HTML]{F8696B}0.349          & \cellcolor[HTML]{F8726C}0.350          & \cellcolor[HTML]{FBA476}0.354          & \cellcolor[HTML]{63BE7B}\textbf{0.367} \\
\textbf{cos-es-char}    & \cellcolor[HTML]{63BE7B}0.819          & \cellcolor[HTML]{FEEA83}0.812          & \cellcolor[HTML]{FFEB84}0.812          & \cellcolor[HTML]{B6D680}0.816          & \cellcolor[HTML]{FCC07B}0.803          & \cellcolor[HTML]{F8696B}0.784          & \cellcolor[HTML]{63BE7B}\textbf{0.819} \\
\textbf{cos-ru-sgns}    & \cellcolor[HTML]{F5E984}0.412          & \cellcolor[HTML]{63BE7B}0.421          & \cellcolor[HTML]{FFEB84}0.411          & \cellcolor[HTML]{FDD37F}0.406          & \cellcolor[HTML]{FCB77A}0.399          & \cellcolor[HTML]{F8696B}0.381          & \cellcolor[HTML]{63BE7B}\textbf{0.421} \\
\textbf{cos-ru-char}    & \cellcolor[HTML]{FEDD81}0.818          & \cellcolor[HTML]{C6DB81}0.822          & \cellcolor[HTML]{FEE582}0.820          & \cellcolor[HTML]{63BE7B}0.824          & \cellcolor[HTML]{F8696B}0.788          & \cellcolor[HTML]{FFEB84}0.821          & \cellcolor[HTML]{63BE7B}\textbf{0.824} \\
\textbf{cos-ru-electra} & \cellcolor[HTML]{63BE7B}0.724          & \cellcolor[HTML]{FFEB84}0.712          & \cellcolor[HTML]{D9E082}0.715          & \cellcolor[HTML]{FEE883}0.712          & \cellcolor[HTML]{F8696B}0.683          & \cellcolor[HTML]{FCBF7B}0.702          & \cellcolor[HTML]{63BE7B}\textbf{0.724} \\
\hline
\textbf{rnk-en-sgns}    & \cellcolor[HTML]{FFEA84}0.247           & \cellcolor[HTML]{81C67C}0.234           & \cellcolor[HTML]{FFEB84}0.246           & \cellcolor[HTML]{63BE7B}0.231           & \cellcolor[HTML]{FDBD7C}0.252            & \cellcolor[HTML]{F8696B}0.262           & \cellcolor[HTML]{63BE7B}\textbf{0.231}  \\
\textbf{rnk-en-char}    & \cellcolor[HTML]{F9E983}0.438           & \cellcolor[HTML]{FFE483}0.439           & \cellcolor[HTML]{63BE7B}0.419           & \cellcolor[HTML]{F8696B}0.448           & \cellcolor[HTML]{FFEB84}0.438            & \cellcolor[HTML]{FCA276}0.444           & \cellcolor[HTML]{63BE7B}\textbf{0.419}  \\
\textbf{rnk-en-electra} & \cellcolor[HTML]{FFEB84}0.438           & \cellcolor[HTML]{F8696B}0.446           & \cellcolor[HTML]{DEE182}0.437           & \cellcolor[HTML]{FA8A72}0.444           & \cellcolor[HTML]{FFEB84}0.438            & \cellcolor[HTML]{63BE7B}0.432           & \cellcolor[HTML]{63BE7B}\textbf{0.432}  \\
\textbf{rnk-it-sgns}    & \cellcolor[HTML]{63BE7B}0.165           & \cellcolor[HTML]{FFEB84}0.177           & \cellcolor[HTML]{FED680}0.178           & \cellcolor[HTML]{9ACE7E}0.169           & \cellcolor[HTML]{F8696B}0.188            & \cellcolor[HTML]{F9766E}0.187           & \cellcolor[HTML]{63BE7B}\textbf{0.165}  \\
\textbf{rnk-it-char}    & \cellcolor[HTML]{FFE583}0.397           & \cellcolor[HTML]{AED37F}0.390           & \cellcolor[HTML]{FB9073}0.400           & \cellcolor[HTML]{F8696B}0.402           & \cellcolor[HTML]{FFEB84}0.397            & \cellcolor[HTML]{63BE7B}0.383           & \cellcolor[HTML]{63BE7B}\textbf{0.383}  \\
\textbf{rnk-fr-sgns}    & \cellcolor[HTML]{FFE583}0.214           & \cellcolor[HTML]{B7D67F}0.203           & \cellcolor[HTML]{FFEB84}0.212           & \cellcolor[HTML]{63BE7B}0.193           & \cellcolor[HTML]{FDBF7C}0.229            & \cellcolor[HTML]{F8696B}0.262           & \cellcolor[HTML]{63BE7B}\textbf{0.193}  \\
\textbf{rnk-fr-char}    & \cellcolor[HTML]{CFDD81}0.425           & \cellcolor[HTML]{FECC7E}0.429           & \cellcolor[HTML]{FFEB84}0.427           & \cellcolor[HTML]{F8696B}0.435           & \cellcolor[HTML]{FCAD78}0.431            & \cellcolor[HTML]{63BE7B}0.421           & \cellcolor[HTML]{63BE7B}\textbf{0.421}  \\
\textbf{rnk-fr-electra} & \cellcolor[HTML]{FFEB84}0.447           & \cellcolor[HTML]{F8696B}0.463           & \cellcolor[HTML]{FFE884}0.448           & \cellcolor[HTML]{FED981}0.450           & \cellcolor[HTML]{E1E282}0.444            & \cellcolor[HTML]{63BE7B}0.429           & \cellcolor[HTML]{63BE7B}\textbf{0.429}  \\
\textbf{rnk-es-sgns}    & \cellcolor[HTML]{63BE7B}0.197           & \cellcolor[HTML]{FA8571}0.214           & \cellcolor[HTML]{FFDE82}0.203           & \cellcolor[HTML]{A6D17E}0.199           & \cellcolor[HTML]{F8696B}0.217            & \cellcolor[HTML]{FFEB84}0.201           & \cellcolor[HTML]{63BE7B}\textbf{0.197}  \\
\textbf{rnk-es-char}    & \cellcolor[HTML]{FFEB84}0.407           & \cellcolor[HTML]{FFDA81}0.409           & \cellcolor[HTML]{63BE7B}0.403           & \cellcolor[HTML]{FDBA7B}0.412           & \cellcolor[HTML]{E1E282}0.407            & \cellcolor[HTML]{F8696B}0.420           & \cellcolor[HTML]{63BE7B}\textbf{0.403}  \\
\textbf{rnk-ru-sgns}    & \cellcolor[HTML]{FDC17C}0.161           & \cellcolor[HTML]{B8D67F}0.153           & \cellcolor[HTML]{F8696B}0.175           & \cellcolor[HTML]{FFEB84}0.154           & \cellcolor[HTML]{FBA176}0.166            & \cellcolor[HTML]{63BE7B}0.150           & \cellcolor[HTML]{63BE7B}\textbf{0.150}  \\
\textbf{rnk-ru-char}    & \cellcolor[HTML]{B3D57F}0.361           & \cellcolor[HTML]{FFEB84}0.365           & \cellcolor[HTML]{FA7F70}0.376           & \cellcolor[HTML]{FBA176}0.372           & \cellcolor[HTML]{F8696B}0.378            & \cellcolor[HTML]{63BE7B}0.357           & \cellcolor[HTML]{63BE7B}\textbf{0.357}  \\
\textbf{rnk-ru-electra} & \cellcolor[HTML]{E1E282}0.350           & \cellcolor[HTML]{FCB179}0.355           & \cellcolor[HTML]{FFEB84}0.351           & \cellcolor[HTML]{F8696B}0.359           & \cellcolor[HTML]{FFE984}0.351            & \cellcolor[HTML]{63BE7B}0.345           & \cellcolor[HTML]{63BE7B}\textbf{0.345}  \\
\hline
\textbf{TOTAL mse}     & \cellcolor[HTML]{63BE7B}\textbf{10.266} & \cellcolor[HTML]{A2D07E}\textbf{10.363} & \cellcolor[HTML]{FFEB84}\textbf{10.504} & \cellcolor[HTML]{FFDF82}\textbf{10.634} & \cellcolor[HTML]{F97A6F}\textbf{11.645}  & \cellcolor[HTML]{F8696B}\textbf{11.817} & \cellcolor[HTML]{63BE7B}\textbf{10.266} \\
\textbf{TOTAL cos}     & \cellcolor[HTML]{63BE7B}\textbf{8.004}  & \cellcolor[HTML]{C3DA81}\textbf{7.971}  & \cellcolor[HTML]{FEE182}\textbf{7.931}  & \cellcolor[HTML]{FFEB84}\textbf{7.951}  & \cellcolor[HTML]{F98770}\textbf{7.742}   & \cellcolor[HTML]{F8696B}\textbf{7.677}  & \cellcolor[HTML]{63BE7B}\textbf{8.004}  \\
\textbf{TOTAL rnk}     & \cellcolor[HTML]{63BE7B}\textbf{4.248}  & \cellcolor[HTML]{FFE984}\textbf{4.276}  & \cellcolor[HTML]{FFEB84}\textbf{4.275}  & \cellcolor[HTML]{D6DF81}\textbf{4.268}  & \cellcolor[HTML]{F8696B}\textbf{4.337}   & \cellcolor[HTML]{FDC57D}\textbf{4.293}  & \cellcolor[HTML]{63BE7B}\textbf{4.248} \\
\hline
\end{tabular}
\caption{\label{tab:rd-test-score-all} Test results for all our \revdict (RD) approaches. For each score, comparative results are shown in color. Green is used for the best and red for the worst-performing solution per row (a metric defines whether higher or lower values are better). The total score is the sum of the values over all datasets and embeddings.}
\end{table*}

\begin{table*}
\centering
\small
\begin{tabular}{l|lll|ll|lll|ll|lll}
\hline
\textbf{TEAM} & \multicolumn{3}{c}{\textbf{EN}} & \multicolumn{2}{c}{\textbf{ES}} & \multicolumn{3}{c}{\textbf{FR}} & \multicolumn{2}{c}{\textbf{IT}} & \multicolumn{3}{c}{\textbf{RU}} \\
\textbf{} & \textbf{sgns} & \textbf{char} & \textbf{electra} & \textbf{sgns} & \textbf{char} & \multicolumn{1}{c}{\textbf{sgns}} & \textbf{char} & \textbf{electra}   & \textbf{sgns} & \textbf{char} & \textbf{sgns} & \textbf{char} & \textbf{electra}   \\
\hline
0                        & \cellcolor[HTML]{FED07F}0.909                             & \cellcolor[HTML]{FFFFFF}                                  & \cellcolor[HTML]{FFFFFF}                                     & \cellcolor[HTML]{F8696B}0.913                             & \cellcolor[HTML]{FFFFFF}                                  & \cellcolor[HTML]{F8696B}1.122                             & \cellcolor[HTML]{FFFFFF}\textbf{}                         & \textbf{}                              & \cellcolor[HTML]{F8696B}1.196          &                                        & \cellcolor[HTML]{F8696B}0.615          &                                        & \textbf{}                              \\
\textbf{1}               & \cellcolor[HTML]{F8696B}\textbf{0.964}                    & \cellcolor[HTML]{FCA377}\textbf{0.162}                    & \cellcolor[HTML]{F8696B}\textbf{1.685}                       & \cellcolor[HTML]{B5D57F}\textbf{0.883}                    & \cellcolor[HTML]{FFEB84}\textbf{0.526}                    & \cellcolor[HTML]{D4DE81}\textbf{1.068}                    & \cellcolor[HTML]{FFEB84}\textbf{0.390}                    & \cellcolor[HTML]{F8696B}\textbf{1.339} & \cellcolor[HTML]{DFE182}\textbf{1.076} & \cellcolor[HTML]{FFE283}\textbf{0.366} & \cellcolor[HTML]{FFE984}\textbf{0.568} & \cellcolor[HTML]{FFEB84}\textbf{0.140} & \cellcolor[HTML]{F8696B}\textbf{0.911} \\
2                        &                                                           &                                                           &                                                              & \cellcolor[HTML]{FB9A75}0.911                             &                                                           &                                                           &                                                           &                                        &                                        &                                        &                                        &                                        &                                        \\
3                    & \cellcolor[HTML]{63BE7B}0.854                             &                                                           &                                                              &                                                           &                                                           &                                                           &                                                           &                                        &                                        &                                        &                                        &                                        &                                        \\
5                         & \cellcolor[HTML]{89C87D}0.864                             & \cellcolor[HTML]{DAE081}0.143                             & \cellcolor[HTML]{90CB7D}1.310                                & \cellcolor[HTML]{69BF7B}0.860                             & \cellcolor[HTML]{63BE7B}0.467                             & \cellcolor[HTML]{63BE7B}1.026                             & \cellcolor[HTML]{63BE7B}0.335                             & \cellcolor[HTML]{63BE7B}1.066          & \cellcolor[HTML]{63BE7B}1.031          & \cellcolor[HTML]{63BE7B}0.334          & \cellcolor[HTML]{88C87D}0.538          & \cellcolor[HTML]{63BE7B}0.116          & \cellcolor[HTML]{69BF7B}0.828          \\
6                                      & \cellcolor[HTML]{FFE383}0.900                             & \cellcolor[HTML]{FFEB84}0.143                             & \cellcolor[HTML]{FFE984}1.340                                &                                                           &                                                           &                                                           &                                                           &                                        &                                        &                                        &                                        &                                        &                                        \\
7                       & \cellcolor[HTML]{FDC67D}0.915                             & \cellcolor[HTML]{FA8B72}0.168                             &                                                              & \cellcolor[HTML]{FFEB84}0.906                             & \cellcolor[HTML]{FCA677}0.557                             & \cellcolor[HTML]{FDB77A}1.100                             & \cellcolor[HTML]{FFE583}0.391                             &                                        & \cellcolor[HTML]{FFE082}1.097          & \cellcolor[HTML]{FFE784}0.364          & \cellcolor[HTML]{FECE7F}0.578          & \cellcolor[HTML]{FDBC7B}0.156          &                                        \\
10                         & \cellcolor[HTML]{B1D47F}0.875                             & \cellcolor[HTML]{63BE7B}0.141                             & \cellcolor[HTML]{63BE7B}1.301                                &                                                           &                                                           &                                                           &                                                           &                                        & \cellcolor[HTML]{FFEB84}1.087          & \cellcolor[HTML]{D8DF81}0.355          &                                        &                                        &                                        \\
12             & \cellcolor[HTML]{FFEB84}0.895                             & \cellcolor[HTML]{E7E482}0.143                             & \cellcolor[HTML]{DCE081}1.326                                & \cellcolor[HTML]{FB9E76}0.910                             & \cellcolor[HTML]{D5DF81}0.510                             & \cellcolor[HTML]{FB9E76}1.107                             & \cellcolor[HTML]{BAD780}0.366                             & \cellcolor[HTML]{9BCE7E}1.112          & \cellcolor[HTML]{FECF7F}1.111          & \cellcolor[HTML]{F0E683}0.359          & \cellcolor[HTML]{FAE983}0.566          & \cellcolor[HTML]{CADB80}0.132          & \cellcolor[HTML]{FEC87E}0.864          \\
13                        & \cellcolor[HTML]{83C77C}0.862                             & \cellcolor[HTML]{F8696B}0.176                             & \cellcolor[HTML]{FCAB78}1.509                                & \cellcolor[HTML]{63BE7B}0.858                             & \cellcolor[HTML]{F8696B}0.583                             & \cellcolor[HTML]{6DC07B}1.030                             & \cellcolor[HTML]{F8696B}0.411                             & \cellcolor[HTML]{FCA577}1.271          & \cellcolor[HTML]{7AC47C}1.039          & \cellcolor[HTML]{F8696B}0.438          & \cellcolor[HTML]{63BE7B}0.528          & \cellcolor[HTML]{F8696B}0.184          & \cellcolor[HTML]{63BE7B}0.828         \\
\hline
\end{tabular}
\caption{\label{tab:rd-test-mse-all-teams} MSE test scores for each team in \revdict task. The results of our team are bold (team 1). For each task, comparative results are shown in color. Green is used for the best and red for the worst-performing solution per column.}
\end{table*}

\begin{table*}
\centering
\small
\begin{tabular}{l|lll|ll|lll|ll|lll}
\hline
\textbf{TEAM} & \multicolumn{3}{c}{\textbf{EN}} & \multicolumn{2}{c}{\textbf{ES}} & \multicolumn{3}{c}{\textbf{FR}} & \multicolumn{2}{c}{\textbf{IT}} & \multicolumn{3}{c}{\textbf{RU}} \\
\textbf{} & \textbf{sgns} & \textbf{char} & \textbf{electra} & \textbf{sgns} & \textbf{char} & \multicolumn{1}{c}{\textbf{sgns}} & \textbf{char} & \textbf{electra}   & \textbf{sgns} & \textbf{char} & \textbf{sgns} & \textbf{char} & \textbf{electra}   \\
\hline
0             & \cellcolor[HTML]{F8696B}0.156          &                                        & \textbf{}                              & \cellcolor[HTML]{F8696B}0.223          &                                        & \cellcolor[HTML]{F8736C}0.216          &                                        & \textbf{}                              & \cellcolor[HTML]{F8696B}-0.004         &                                        & \cellcolor[HTML]{F8696B}0.006          &                                        & \textbf{}                              \\
\textbf{1}    & \cellcolor[HTML]{63BE7B}\textbf{0.260} & \cellcolor[HTML]{F8696B}\textbf{0.770} & \cellcolor[HTML]{F8696B}\textbf{0.828} & \cellcolor[HTML]{C6DB81}\textbf{0.367} & \cellcolor[HTML]{F8696B}\textbf{0.819} & \cellcolor[HTML]{63BE7B}\textbf{0.342} & \cellcolor[HTML]{F98C71}\textbf{0.756} & \cellcolor[HTML]{F8696B}\textbf{0.847} & \cellcolor[HTML]{63BE7B}\textbf{0.380} & \cellcolor[HTML]{FED980}\textbf{0.724} & \cellcolor[HTML]{6BC17C}\textbf{0.421} & \cellcolor[HTML]{FED980}\textbf{0.824} & \cellcolor[HTML]{FA9874}\textbf{0.724} \\
2             &                                        &                                        &                                        & \cellcolor[HTML]{63BE7B}0.403          &                                        &                                        &                                        &                                        &                                        &                                        &                                        &                                        &                                        \\
3             & \cellcolor[HTML]{85C87D}0.248          &                                        &                                        &                                        &                                        &                                        &                                        &                                        &                                        &                                        &                                        &                                        &                                        \\
5             & \cellcolor[HTML]{99CE7F}0.241          & \cellcolor[HTML]{FAEA84}0.795          & \cellcolor[HTML]{63BE7B}0.847          & \cellcolor[HTML]{FFEB84}0.347          & \cellcolor[HTML]{63BE7B}0.839          & \cellcolor[HTML]{A4D17F}0.312          & \cellcolor[HTML]{63BE7B}0.789          & \cellcolor[HTML]{63BE7B}0.862          & \cellcolor[HTML]{6DC17C}0.374          & \cellcolor[HTML]{63BE7B}0.747          & \cellcolor[HTML]{C7DB81}0.383          & \cellcolor[HTML]{63BE7B}0.852          & \cellcolor[HTML]{63BE7B}0.735          \\
6             & \cellcolor[HTML]{FCB77A}0.185          & \cellcolor[HTML]{E9E583}0.796          & \cellcolor[HTML]{B3D680}0.846          &                                        &                                        &                                        &                                        &                                        &                                        &                                        &                                        &                                        &                                        \\
7             & \cellcolor[HTML]{FDCE7E}0.194          & \cellcolor[HTML]{FEDC81}0.792          &                                        & \cellcolor[HTML]{FA9172}0.262          & \cellcolor[HTML]{F8726C}0.820          & \cellcolor[HTML]{F98C71}0.228          & \cellcolor[HTML]{FFEB84}0.769          &                                        & \cellcolor[HTML]{FEE482}0.260          & \cellcolor[HTML]{B0D580}0.739          & \cellcolor[HTML]{FEE282}0.335          & \cellcolor[HTML]{D0DE82}0.836          &                                        \\
10            & \cellcolor[HTML]{FFEB84}0.204          & \cellcolor[HTML]{63BE7B}0.798          & \cellcolor[HTML]{FEDB80}0.843          &                                        &                                        &                                        &                                        &                                        & \cellcolor[HTML]{FFEB84}0.274          & \cellcolor[HTML]{E5E483}0.734          &                                        &                                        &                                        \\
12            & \cellcolor[HTML]{F9826F}0.166          & \cellcolor[HTML]{FFEB84}0.795          & \cellcolor[HTML]{FEE382}0.844          & \cellcolor[HTML]{F98770}0.252          & \cellcolor[HTML]{FBEA84}0.824          & \cellcolor[HTML]{F8696B}0.212          & \cellcolor[HTML]{F6E984}0.770          & \cellcolor[HTML]{FEE883}0.858          & \cellcolor[HTML]{FEDD81}0.246          & \cellcolor[HTML]{FEE382}0.728          & \cellcolor[HTML]{FDD47F}0.298          & \cellcolor[HTML]{FFEB84}0.830          & \cellcolor[HTML]{F8696B}0.721          \\
13            & \cellcolor[HTML]{92CC7E}0.243          & \cellcolor[HTML]{FBA676}0.782          & \cellcolor[HTML]{9ECF7F}0.846          & \cellcolor[HTML]{EEE784}0.353          & \cellcolor[HTML]{FFEB84}0.824          & \cellcolor[HTML]{81C77D}0.328          & \cellcolor[HTML]{F8696B}0.752          & \cellcolor[HTML]{F3E884}0.859          & \cellcolor[HTML]{82C77D}0.360          & \cellcolor[HTML]{F8696B}0.681          & \cellcolor[HTML]{63BE7B}0.424          & \cellcolor[HTML]{F8696B}0.791          & \cellcolor[HTML]{81C77D}0.734         \\

\hline
\end{tabular}
\caption{\label{tab:rd-test-cos-all-teams} COS test scores for each team in \revdict task. The results of our team are bold (team 1). For each task, comparative results are shown in color. Green is used for the best and red for the worst-performing solution per column.}
\end{table*}

\begin{table*}
\centering
\small
\begin{tabular}{l|lll|ll|lll|ll|lll}
\hline
\textbf{TEAM} & \multicolumn{3}{c}{\textbf{EN}} & \multicolumn{2}{c}{\textbf{ES}} & \multicolumn{3}{c}{\textbf{FR}} & \multicolumn{2}{c}{\textbf{IT}} & \multicolumn{3}{c}{\textbf{RU}} \\
\textbf{} & \textbf{sgns} & \textbf{char} & \textbf{electra} & \textbf{sgns} & \textbf{char} & \multicolumn{1}{c}{\textbf{sgns}} & \textbf{char} & \textbf{electra} & \textbf{sgns} & \textbf{char} & \textbf{sgns} & \textbf{char} & \textbf{electra}   \\
\hline
0             & \cellcolor[HTML]{F96B6C}0.499          &                                        & \textbf{}                              & \cellcolor[HTML]{F8696B}0.495          &                                        & \cellcolor[HTML]{F8696B}0.498          &                                        & \textbf{}                              & \cellcolor[HTML]{F8696B}0.499          &                                        & \cellcolor[HTML]{F8696B}0.499          &                                        & \textbf{}                              \\
\textbf{1}    & \cellcolor[HTML]{63BE7B}\textbf{0.231} & \cellcolor[HTML]{63BE7B}\textbf{0.419} & \cellcolor[HTML]{63BE7B}\textbf{0.432} & \cellcolor[HTML]{99CD7E}\textbf{0.197} & \cellcolor[HTML]{63BE7B}\textbf{0.403} & \cellcolor[HTML]{63BE7B}\textbf{0.193} & \cellcolor[HTML]{A6D17E}\textbf{0.421} & \cellcolor[HTML]{63BE7B}\textbf{0.429} & \cellcolor[HTML]{63BE7B}\textbf{0.165} & \cellcolor[HTML]{63BE7B}\textbf{0.383} & \cellcolor[HTML]{63BE7B}\textbf{0.150} & \cellcolor[HTML]{63BE7B}\textbf{0.357} & \cellcolor[HTML]{63BE7B}\textbf{0.345} \\
2             &                                        &                                        &                                        & \cellcolor[HTML]{63BE7B}0.167          &                                        &                                        &                                        &                                        &                                        &                                        &                                        &                                        &                                        \\
3             & \cellcolor[HTML]{EEE683}0.319          &                                        &                                        &                                        &                                        &                                        &                                        &                                        &                                        &                                        &                                        &                                        &                                        \\
5             & \cellcolor[HTML]{FAE983}0.326          & \cellcolor[HTML]{F96F6D}0.500          & \cellcolor[HTML]{FCA276}0.490          & \cellcolor[HTML]{FFE283}0.271          & \cellcolor[HTML]{FED981}0.424          & \cellcolor[HTML]{F7E883}0.302          & \cellcolor[HTML]{FFEB84}0.428          & \cellcolor[HTML]{F9776E}0.476          & \cellcolor[HTML]{9FCF7E}0.197          & \cellcolor[HTML]{F0E683}0.428          & \cellcolor[HTML]{E2E282}0.247          & \cellcolor[HTML]{FFEB84}0.389          & \cellcolor[HTML]{FA8A72}0.417          \\
6             & \cellcolor[HTML]{F8696B}0.500          & \cellcolor[HTML]{F8696B}0.500          & \cellcolor[HTML]{F8696B}0.500          &                                        &                                        &                                        &                                        &                                        &                                        &                                        &                                        &                                        &                                        \\
7             & \cellcolor[HTML]{FECA7E}0.374          & \cellcolor[HTML]{F2E783}0.478          &                                        & \cellcolor[HTML]{FCAA78}0.375          & \cellcolor[HTML]{DDE182}0.410          & \cellcolor[HTML]{FB9273}0.439          & \cellcolor[HTML]{63BE7B}0.416          &                                        & \cellcolor[HTML]{FCA577}0.384          & \cellcolor[HTML]{FFE283}0.438          & \cellcolor[HTML]{FFDF82}0.291          & \cellcolor[HTML]{C4DA80}0.377          &                                        \\
10            & \cellcolor[HTML]{FDBA7B}0.394          & \cellcolor[HTML]{FFEB84}0.483          & \cellcolor[HTML]{FFEB84}0.478          &                                        &                                        &                                        &                                        &                                        & \cellcolor[HTML]{FCA477}0.386          & \cellcolor[HTML]{FA8E73}0.478          &                                        &                                        &                                        \\
12            & \cellcolor[HTML]{E3E382}0.312          & \cellcolor[HTML]{ADD37F}0.450          & \cellcolor[HTML]{6BC07B}0.434          & \cellcolor[HTML]{FFEB84}0.253          & \cellcolor[HTML]{FFEB84}0.412          & \cellcolor[HTML]{FFE884}0.314          & \cellcolor[HTML]{FFEB84}0.428          & \cellcolor[HTML]{A5D17E}0.442          & \cellcolor[HTML]{FFEB84}0.247          & \cellcolor[HTML]{CCDC81}0.417          & \cellcolor[HTML]{FFDF82}0.290          & \cellcolor[HTML]{FECC7E}0.410          & \cellcolor[HTML]{E8E482}0.399          \\
13            & \cellcolor[HTML]{FFEB84}0.329          & \cellcolor[HTML]{FED781}0.486          & \cellcolor[HTML]{FEEA83}0.478          & \cellcolor[HTML]{FAE983}0.251          & \cellcolor[HTML]{F8696B}0.500          & \cellcolor[HTML]{DCE081}0.282          & \cellcolor[HTML]{F8696B}0.502          & \cellcolor[HTML]{F8696B}0.478          & \cellcolor[HTML]{DEE182}0.230          & \cellcolor[HTML]{F8696B}0.496          & \cellcolor[HTML]{93CC7D}0.187          & \cellcolor[HTML]{F8696B}0.472          & \cellcolor[HTML]{F8696B}0.420         \\
\hline
\end{tabular}
\caption{\label{tab:rd-test-rnk-all-teams} RNK test scores for each team in \revdict task. The results of our team are bold (team 1). For each task, comparative results are shown in color. Green is used for the best and red for the worst performing solution per column.}
\end{table*}

\end{document}